\documentclass[conference]{IEEEtran}
\IEEEoverridecommandlockouts
\usepackage{cite}
\usepackage{amsmath,amssymb,amsfonts}
\usepackage{textcomp}
\def\BibTeX{{\rm B\kern-.05em{\sc i\kern-.025em b}\kern-.08em
    T\kern-.1667em\lower.7ex\hbox{E}\kern-.125emX}}

\usepackage[T1]{fontenc}    % use 8-bit T1 fonts
\usepackage[utf8]{inputenc} % allow utf-8 input

\usepackage{algorithm}
\usepackage{algpseudocode}

\usepackage{amsfonts}       % blackboard math symbols
\usepackage{graphicx}
\usepackage{hyperref}       % hyperlinks
\usepackage{booktabs}       % professional-quality tables
\usepackage{listings}
\usepackage{longtable}
\usepackage{nicefrac}       % compact symbols for 1/2, etc.
\usepackage{microtype}      % microtypography
\usepackage{multirow}
\usepackage{subcaption}
\usepackage{tabularx}
\usepackage{tikz}
\usepackage{todonotes}
\usepackage{url}            % simple URL typesetting
\usepackage{xcolor}         % colors
\usepackage{xspace}

\definecolor{jaxblue}{HTML}{4A90D9}
\definecolor{cppred}{HTML}{D94A4A}
\definecolor{jsyellow}{HTML}{E8A838}
\definecolor{rlgreen}{HTML}{3DAA5C}
\definecolor{searchpurple}{HTML}{8B5CF6}
\definecolor{evogold}{HTML}{D4A017}
\definecolor{llmteal}{HTML}{2CAAA8}
\definecolor{jitbg}{HTML}{E8F0FE}
\definecolor{backendgray}{HTML}{F0F0F0}
\definecolor{dataorange}{HTML}{E07020}
\definecolor{validgreen}{HTML}{2E8B57}

\newcommand{\name}{\textit{PuzzleJAX}\xspace}
\newcommand{\puzzlescript}{{PuzzleScript}\xspace}
\newcommand{\code}[1]{\texttt{#1}}
\newcommand{\game}[1]{\textit{\textbf{#1}}}

\definecolor{ps-brown}{rgb}{0.65, 0.16, 0.16}
\definecolor{ps-orange}{rgb}{1.0, 0.65, 0}
\definecolor{ps-white}{rgb}{1, 1, 1}
\definecolor{ps-blue}{rgb}{0, 0, 1}
\definecolor{ps-commentblue}{rgb}{0.13725490196078433, 0.4549019607843137, 0.8666666666666667}
\definecolor{ps-yellow}{rgb}{1, 1, 0}
\definecolor{ps-darkred}{rgb}{0.55, 0, 0}
\definecolor{ps-red}{rgb}{1, 0, 0}
\definecolor{ps-lightred}{rgb}{1, 0.41, 0.41}
\definecolor{ps-green}{rgb}{0, 1, 0}
\definecolor{ps-black}{rgb}{0, 0, 0}
\definecolor{ps-keywordpurple}{rgb}{0.564, 0.427, 0.843}
\definecolor{ps-purple}{rgb}{0.6901960784313725, 0.11372549019607843, 0.7568627450980392} 
\definecolor{ps-background}{rgb}{0.058823529411764705, 0.09803921568627451, 0.16470588235294117}
\lstdefinestyle{puzzlescript}{
    backgroundcolor=\color{ps-background},   
    basicstyle=\ttfamily\footnotesize\color{white},
    keywordstyle={\color{ps-keywordpurple}},
    commentstyle=\color{ps-commentblue},
    stringstyle=\color{ps-orange},
    numbers=none,
    frame=none,
    framerule=0pt,
    framexleftmargin=0em,
    framexrightmargin=0em,
    framextopmargin=0pt,
    framexbottommargin=0pt,
    xleftmargin=0em,
    xrightmargin=0em,
    breaklines=false,
    breakatwhitespace=false,
    showspaces=false,
    showstringspaces=false,
    showtabs=false,
    columns=flexible,
    keepspaces=true,
    moredelim=[is][\color{gray}]{startlevel}{endlevel},
    morecomment=[s]{(}{)},
    literate={->}{{{\color{ps-red}{->}}}}2
             {>}{{{\color{ps-purple}{>}}}}1
             {<}{{{\color{ps-purple}{<}}}}1
             {\\wedge}{{\color{ps-purple}{$\wedge${}}}}3
             {\\vee}{{{\color{ps-purple}{$\vee$}}}}4,
    keywords={title, author, OBJECTS, LEGEND, SOUNDS, COLLISIONLAYERS, RULES, WINCONDITIONS, all, on, LEVELS},
    keywords=[2]{Background, Player, Box, Trigger, Switch, GateClosed, GateOpen, Wall, Toggle},
    keywordstyle=[2]{\color{ps-green}},
    keywords=[3]{Unconventional, PushPull, ChatGPT},
    keywordstyle=[3]{\color{ps-orange}},
    keywords=[4]{late, no},
    keywordstyle=[4]{\color{ps-purple}}
}

\begin{document}

\title{PuzzleJAX: A Benchmark for Reasoning and Learning}

\author{
\IEEEauthorblockN{   Sam Earle$^1$ \quad Graham Todd$^1$ \quad Yuchen Li$^1$ \quad Ahmed Khalifa$^2$ \quad Zehua Jiang$^1$\\
Muhammad Umair Nasir$^3$ \quad Andrzej Banburski-Fahey$^4$ \quad Julian Togelius$^1$}
\IEEEauthorblockA{$^1$\textit{New York University} \quad $^2$\textit{University of Malta}\\
$^3$\textit{University of the Witwatersrand} \quad $^4$\textit{Microsoft}\\
sam.earle@nyu.edu}  
% \texttt{\{sam.earle,gdrtodd,yl6394\}}@nyu.edu \quad \texttt{ahmed@akhalifa.com}\\
% \texttt{zj2086@nyu.edu} \quad \texttt{muhammad.nasir@wits.ac.za} \\
% \texttt{abanburski@microsoft.com} \quad \texttt{julian@togelius.com}
}

% \author{anon.}

\maketitle

\begin{abstract}

% We're intested in whether game-playing AI agents are capable of creative insight in .
We're interested in developing tools that surface the question of creative insight in game-playing AI.
\puzzlescript is a popular open-source grid-based game engine which has facilitated the design of a number of high-quality lateral thinking puzzles.
To make this growing space of interesting problems amenable to AI research, we introduce the \name framework, a Python interface for \puzzlescript, with ground truth (JavaScript), multi-threaded (C++), and GPU-accelerated (JAX) backends.
We include scripts for AI player agents including tree search, reinforcement learning, and large language models.
Preliminary benchmarking on a select set of games shows that simple search is surprisingly competitive compared to reinforcement learning and large language models.\footnote{Our code is available at \url{https://github.com/smearle/script-doctor}}
% We additionally collect a dataset of thousands of human games published via the engine since its release in 2013.
% We collect over $26,000$ unique PuzzleScript games,
% , thereby demonstrating \name's coverage of an expansive, expressive, and human-relevant space of tasks.
% By analyzing the performance of search, learning, and large language models on these games, we show that \name can naturally express tasks that are both simple and intuitive to understand, yet often deeply challenging to master, requiring a combination of control, planning, and creative insight.
% We conclude by setting forth a roadmap for future progress on the \name benchmark, and discuss possible hybrid algorithmic approaches that may be crucial in achieving these goals.
\end{abstract}

\section{Introduction}

Games---from board games to card games to video games---have long been used to train and test methods in artificial intelligence (AI). While ``classic'' game-AI research has largely focused on search and planning (i.e. for superhuman play of traditional board games \cite{tesauro1995temporal,  campbell2002deep, schaeffer2007checkers, silver2016mastering, Silver_2018_AlphaZero}), games as a whole are diverse enough to test a wide variety of cognitive skills. In recent years, specialized game-based benchmarks have been developed to test the capabilities of AI systems in a variety of domains \cite{cui2025tales, nasir2024gametraversalbenchmark, bailis2024werewolf, yannakakis2025artificial}.

% Games, including board game, card games, and various types of video games, have been used to train and test AI methods for a long time. The beauty of this is that depending on the particular game, and how it is represented to the AI system, it can test different AI capabilities. This includes learning, planning, and reasoning; specialized game-based benchmarks have been developed for different methods, such as tree search, reinforcement learning, and large language models~\cite{yannakakis2025artificial}. 

Relative to other genres (e.g. strategy games, platforming games, arcade games), \textit{puzzle games} have received comparatively less research attention. These games are typically single-player, with full or nearly full state observability and relatively modest action spaces. What puzzle games lack in dexterity-based challenges, they make up for in tests of logical inference and long-horizon planning. Puzzle games also range in the complexity of their observation space from relatively simple (e.g. the tile-based levels of \textit{Sokoban}, \textit{Boulder Dash}, or \textit{Baba is You}) to expansive and immersive (e.g. the fully-realized 3D worlds of \textit{Portal}, \textit{The Witness}, or \textit{The Talos Principle}). We argue that even simple tile-based puzzle games represent an important unsolved frontier in game AI research and help test increasingly important aspects of artificial ``cognition'' in the era of large language models.

\begin{figure}
\centering
\includegraphics[width=\linewidth,trim=220 15 200 50,clip]{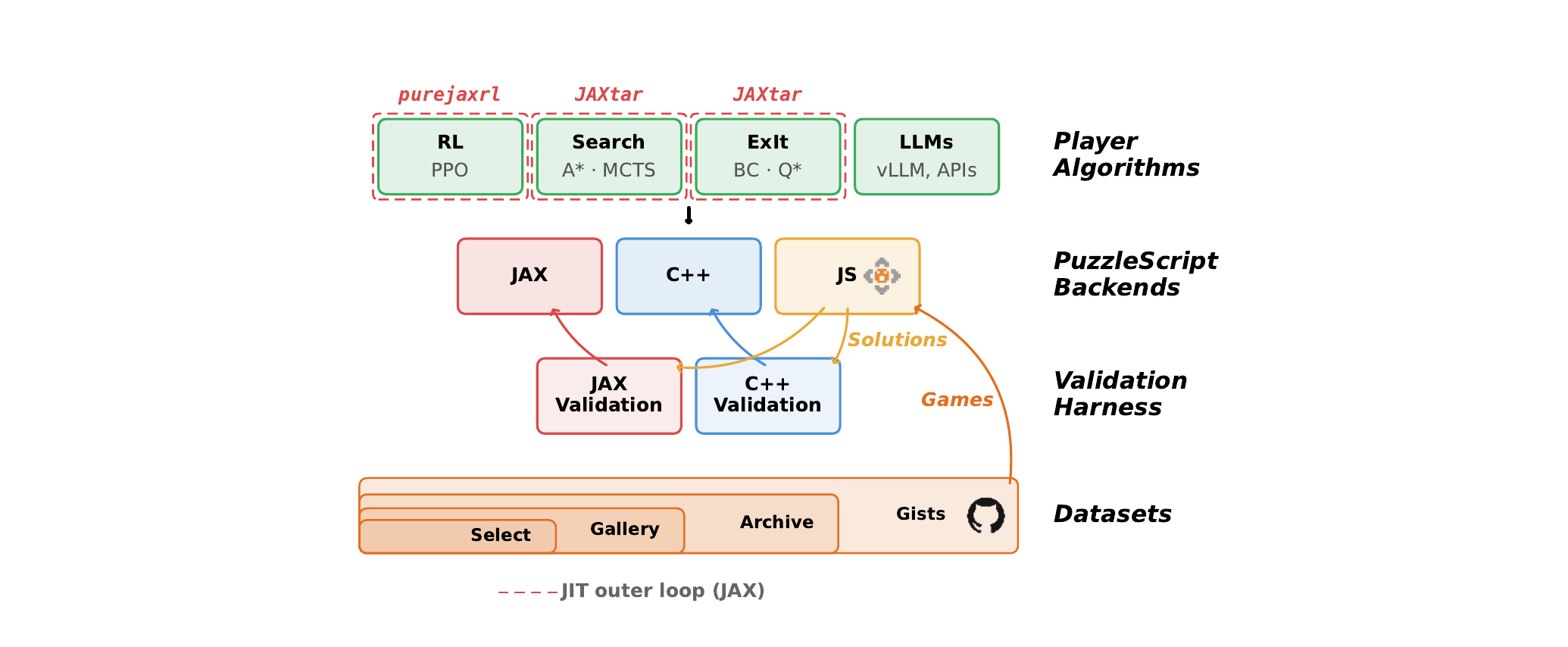}
\caption{\textbf{The \name framework} facilitates the use of AI agents, search, and optimization for playing \puzzlescript games. A dataset of over $26,000$ game files is collected via public gists. Search is used to generate (partial) solutions to valid games in the original JavaScript engine, and a validation harness ensures functional equivalency of multi-threaded/GPU-accelerated reimplementations in C++ and JAX.
}
\label{fig:frontfig}
\end{figure}

% A genre of games that have seen less interest than many others are puzzle video games. These are typically single-player games, with full or mostly full information, and typically discrete game state and action space. What these games lack in twitch-based challenges, they make up for in puzzle challenges. They typically revolve around relatively long-horizon planning, where movement in game space and judicious use of the action space are used to bring about a reconfiguration of the game world. Classic puzzle games include Sokoban, Boulder Dash, and Lemmings, while modern players might be better aquainted with games such as The Witness, The Talos Principle, and Baba is You. Separately, there are matching tile puzzles such as Tetris, Candy Crush Saga, and Lumines.  

Rather than isolating a single puzzle game or group of games as a target or benchmark, we propose a framework for analyzing and evaluating tile-based puzzle games more generally. Our approach builds on \puzzlescript, a popular domain-specific language for expressing 2D tile-based puzzle games. We reimplement the \puzzlescript game engine in C++ and JAX, a modern Python library for hardware-accelerated code. The end result is a benchmark of over 2,500 distinct game environments and the capacity to generate and automatically compile completely novel rulesets. Our benchmark, \name, avoids the common problem of model overfitting by offering a vast array of environment dynamics and objectives while still providing a unified observation and action space. \name is completely interoperable with existing \puzzlescript game descriptions, giving easy access to thousands of unique and human-authored game environments. \name is also fast: by leveraging the power of modern computing hardware, we achieve speed-ups in all the tested games ranging from $2\times$ to $16\times$ compared to existing implementations in JavaScript. 
% Finally, \name provides a convenient and familiar environment API, facilitating the use of existing deep learning frameworks with an entire range of novel environments.

% \begin{subfigure}{\linewidth}
% \centering
% \includegraphics[width=0.19\textwidth]{figs/take_heart_lass/level-9_frame-im-0.png}
% \includegraphics[width=0.19\textwidth]{figs/take_heart_lass/level-9_frame-im-5.png}
% \includegraphics[width=0.19\textwidth]{figs/take_heart_lass/level-9_frame-im-7.png}
% \includegraphics[width=0.19\textwidth]{figs/take_heart_lass/level-9_frame-im-11.png}
% \includegraphics[width=0.19\textwidth]{figs/take_heart_lass/level-9_frame-im-14.png}
% \caption{In \textbf{Take Heart Lass}, the player must reach the exit (red heart) before they are blocked by the gradually-spreading despair (black tiles). They can push pink hearts to block the despair or unblock hope (pink tiles) which spreads and consumes despair.}
% \label{fig:take_heart_lass}
% \end{subfigure}

\begin{figure}
\centering
\begin{subfigure}{\linewidth}
\centering
\includegraphics[width=0.19\textwidth]{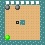}
\includegraphics[width=0.19\textwidth]{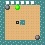}
\includegraphics[width=0.19\textwidth]{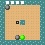}
\includegraphics[width=0.19\textwidth]{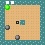}
\includegraphics[width=0.19\textwidth]{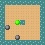}
\caption{In \textbf{Slidings}, the player can swap between boulders (grey circles, light green when selected) and ``slide'' them in any of the cardinal directions until they hit a wall or another boulder. The player has limited moves (top row of white squares). They must arrange boulders to set up future shots and ultimately slide a boulder to a stationary spot over a target (dark blue hollow square).}
\label{fig:slidings}
\end{subfigure}

\begin{subfigure}{\linewidth}
\centering
\includegraphics[width=0.19\linewidth]{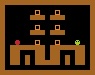}
\includegraphics[width=0.19\linewidth]{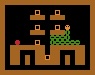}
\includegraphics[width=0.19\linewidth]{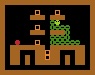}
\includegraphics[width=0.19\linewidth]{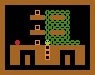}
\includegraphics[width=0.19\linewidth]{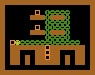}
\caption{In \textbf{Lime Rick}, the player controls a caterpillar-like creature whose head can rise vertically by at most 3 tiles consecutively. The player must navigate the level using their own body and pushable crates to fight against gravity and reach the exit.}
\label{fig:limerick}
\end{subfigure}

\begin{subfigure}{\linewidth}
\centering
\includegraphics[width=0.19\textwidth]{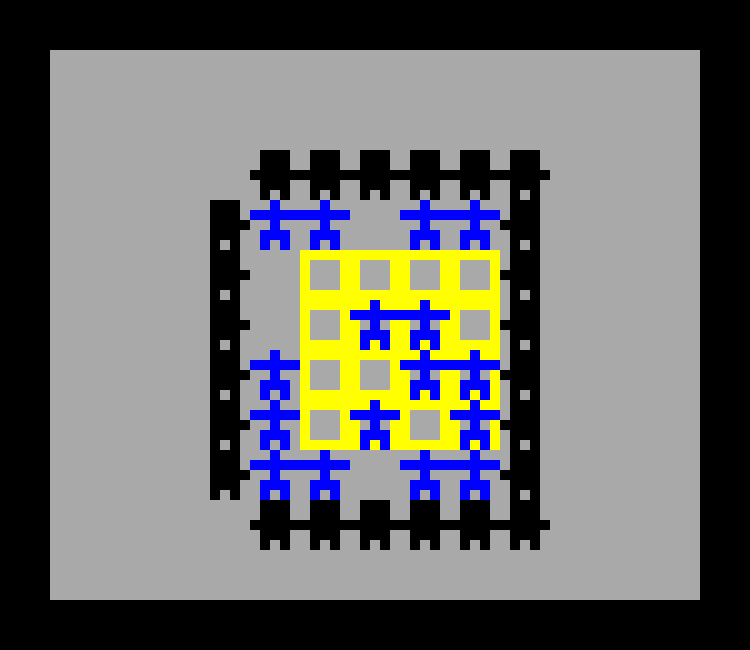}
\includegraphics[width=0.19\textwidth]{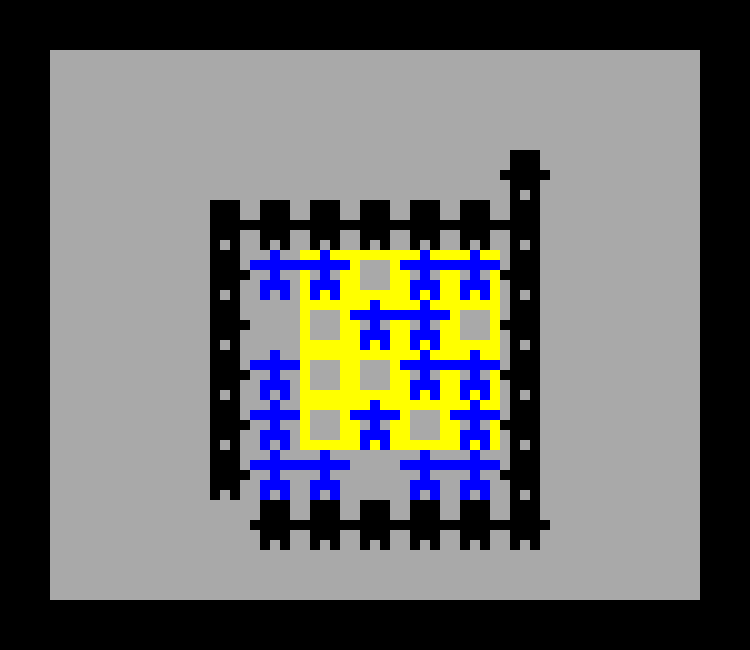}
\includegraphics[width=0.19\textwidth]{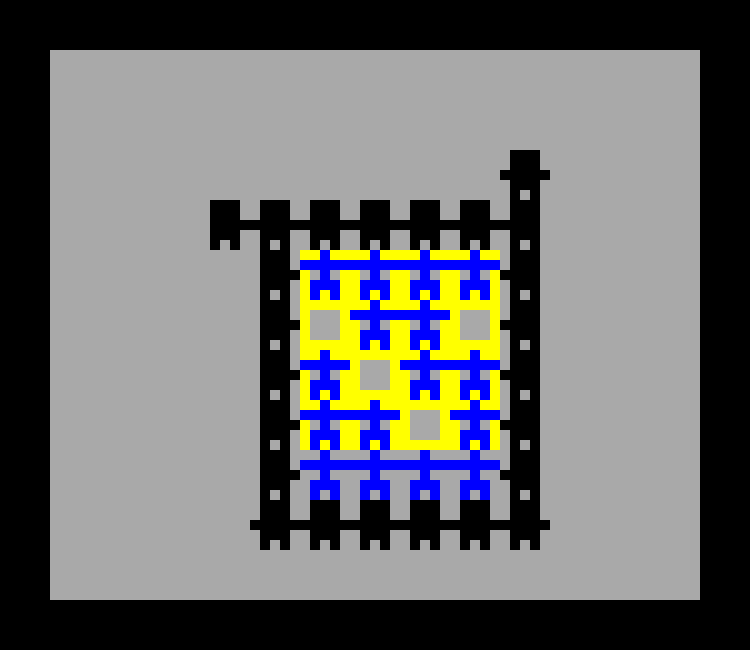}
\includegraphics[width=0.19\textwidth]{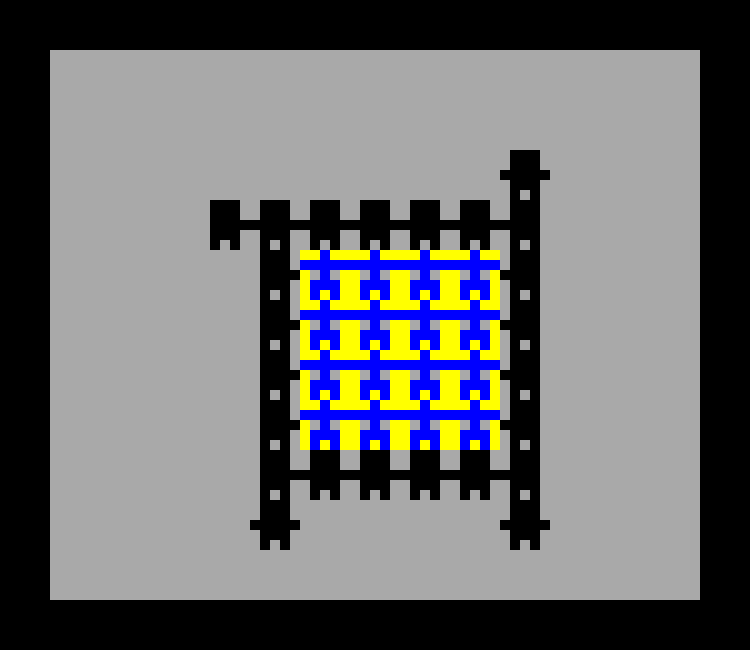}
\includegraphics[width=0.19\textwidth]{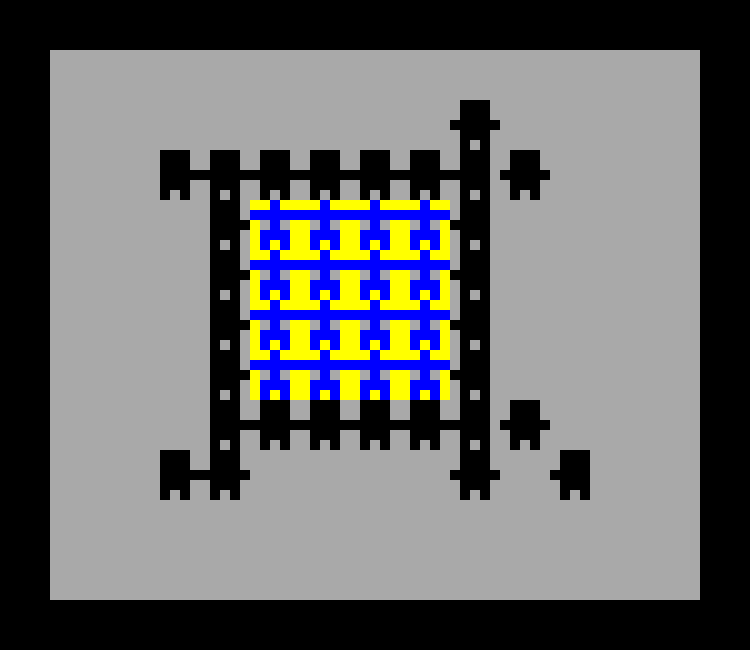}
\caption{In \textbf{Kettle}, the player controls multiple walls of policemen---each of which can move in one direction together---and must strategically sequence moves to push (or ``kettle'') a group of civilians into a compact, confined square.}
\label{fig:kettle}
\end{subfigure}
\caption{Example games from the framework that Select dataset, showcasing the diversity of \puzzlescript games.}
\label{fig:example_games}
\end{figure}

In the following sections, we describe the \name language and implementation, provide comparisons to the existing \puzzlescript implementation, and showcase initial examples of planning algorithms, reinforcement learning, and LLM-based players interacting with puzzle game environments.
Preliminary benchmarking results on a subset of human-authored games demonstrate that \name environments often present substantial challenges for LLM and RL player agents despite being relatively easy to solve via tree search and tractable for human players.

\section{Related work}

% Individual games have been a test bed for AI algorithms, especially Reinforcement Learning algorithms~\cite{szita2012reinforcement}, for many years.
In AI research, individual board or video games are often used to benchmark algorithms like Tree Search~\cite{coulom2006efficient,silver2010monte,browne2012survey} and Reinforcement Learning~\cite{szita2012reinforcement}.
% This is because games are well-defined, self-contained problems that can nonetheless offer significant complexity to players.
% Moreover, the most popular games have stood the test of time by providing relevant and entertaining challenges to human players.
% Because they are accessible and salient to the general public, showcasing AI abilities in such games is an effective means of conveying these agents' capabilities.
% The reason behind this is the complexity a game offers to an AI algorithm, which can help in benchmarking planning, reasoning, and learning.
% For example, AlphaGO~\cite{silver2016mastering} was an agent that learned to play \textit{Go}, which defeated the world champion in the game.
% These capabilities are especially obvious when AI agents can defeat human experts in multiplayer games.
Notably, the ancient board game \textit{Go} was tackled by the tailor-made algorithm AlphaGO~\cite{silver2016mastering}, which combined imitation learning, tree search, and reinforcement learning to defeat world champion Lee Sedol.
Similarly, AlphaStar~\cite{vinyals2019grandmaster} defeated professional players in StarCraft 2, a game known to be one of the most challenging real-time strategy games, and OpenAI Five~\cite{berner2019dota} defeated professional Dota 2 players. 
% Hence, games have been stepping stones for researchers in demonstrating progress on AI algorithms. To this extent, previous works have seen many games turning into AI benchmarks. 
Single player video games, for their part, can also serve as impactful benchmarks, with AI progress reflected incrementally in terms of increasing score or other metrics of in-game progress.
The Arcade Learning Environment~\cite{bellemare2013arcade}, which emulates Atari 2600 games, spurned seminal progress in deep RL~\cite{mnih2013playing}.
The classic platformer \textit{Super Mario Bros.} has similarly served as a benchmark for AI players~\cite{firoiu2017beating, khambhayata2024mastering, ortega2013imitating}.
\textit{Minecraft}~\cite{duncan2011minecraft}, a popular 3D open-world game, has been used as a benchmark for planning and learning in RL agents in a significantly more open-ended task space~\cite{baai2023plan4mc, oh2016control}. 
% \cite{nichol2018gotta} introduced Sonic The Hedgehog as an environment RL agents. 
% Last but not least, 

Beyond playing individual games, \textit{general} game-playing---involving player agents that can play a variety of games or generalize to new environments after learning---has been a core interest among RL researchers. The General Video Game AI (GVGAI)~\cite{perez2019general} research effort leveraged the Video Game Description Language (VGDL)~\cite{ebner2013towards} a Domain Specific Language (DSL) designed to support a large set of arcade-style games, and studied the problem of generalization in RL~\cite{torrado2018deep, justesen2018illuminating, ojha2021cross}. Similarly, the NetHack Learning Environment~\cite{kuttler2020nethack} (a port of NetHack) and Crafter~\cite{hafner2021benchmarking} (a 2D version of Minecraft) were developed to benchmark generalisation in RL algorithms, with their focus on procedural generation prohibiting learning methods prone to overfitting. \name follows in this line of work, supporting thousands of existing human games while also providing a DSL that is capable of expressing a diverse range of game mechanics.
To the best of our knowledge, \name is the first JAX-compatible DSL for grid-based puzzle games.

Due to the high sample complexity of RL algorithms, a growing body of work has used JAX (a GPU-accelerated language) to implement environments, speeding up training both by accelerating environment simulation itself and removing the need for communication between CPU-bound environments and GPU-bound neural controllers.
Numerous suites of JAX-accelerated environments have been introduced, including combinatorial puzzles~\cite{bonnet2024jumanji}, Atari-like arcade games and continuous control tasks~\cite{gymnax2022github}, and two-player board games~\cite{koyamada2023pgx}.
Generally, porting complex, open-ended game environments to JAX involves substantial engineering overhead due to its requirement for fixed-size arrays. 
Nonetheless, such efforts have been undertaken with Craftax~\cite{matthews2024craftax} (an accelerated version of Crafter incorporating additional features inspired by Nethack) and XLand-minigrid~\cite{nikulin2024xland} (a simplified version of XLand~\cite{team2021open} in a minigrid-style environment~\cite{chevalier2023minigrid}).
Like these prior environments, which utilize procedural environment layouts and dynamics to provide a dynamic set of challenges for learning agents, \name is also intended to support open-ended reinforcement learning.
But where existing work has relatively fixed semantics, \name allows for the efficient exploration of a rich space of semantically novel games.
% JAX  been used to explore Unsupervised Environment Design, a high-sample-complexity RL pipeline in which environments are procedurall generated to provide optimal learning challenges for agents.
% JAX is mostly used to implement problems outside of games such as Kinetix~\cite{matthews2024kinetix}, a physics-based environment for control tasks. 

\name can also be seen as a successor in a line of work seeking to benchmark the planning and reasoning abilities of Large Language Models (LLMs) and Vision Language Models (VLMs).
% Previously, GameTraversalBenchmark~\cite{nasir2024gametraversalbenchmark} created a procedurally generated 2D games where LLMs were benchmarked for planning abilities by traversing the maps.
SmartPlay~\cite{wu2023smartplay} tested LLMs on a benchmark of 6 games including Minecraft and Crafter. Dsgbench~\cite{tang2025dsgbench} used 6 strategy games to assess decision-making abilities in LLMs.
Similarly, Balrog~\cite{paglieri2025balrog} introduces a benchmark consisting of 6 learning environments, including Crafter and the NetHack Learning Environment, for testing agentic capabilities of long-context LLMs and VLMs.

\begin{figure*}[t]
\centering
\includegraphics[width=1.0\linewidth, trim=0 0 0 0, clip]{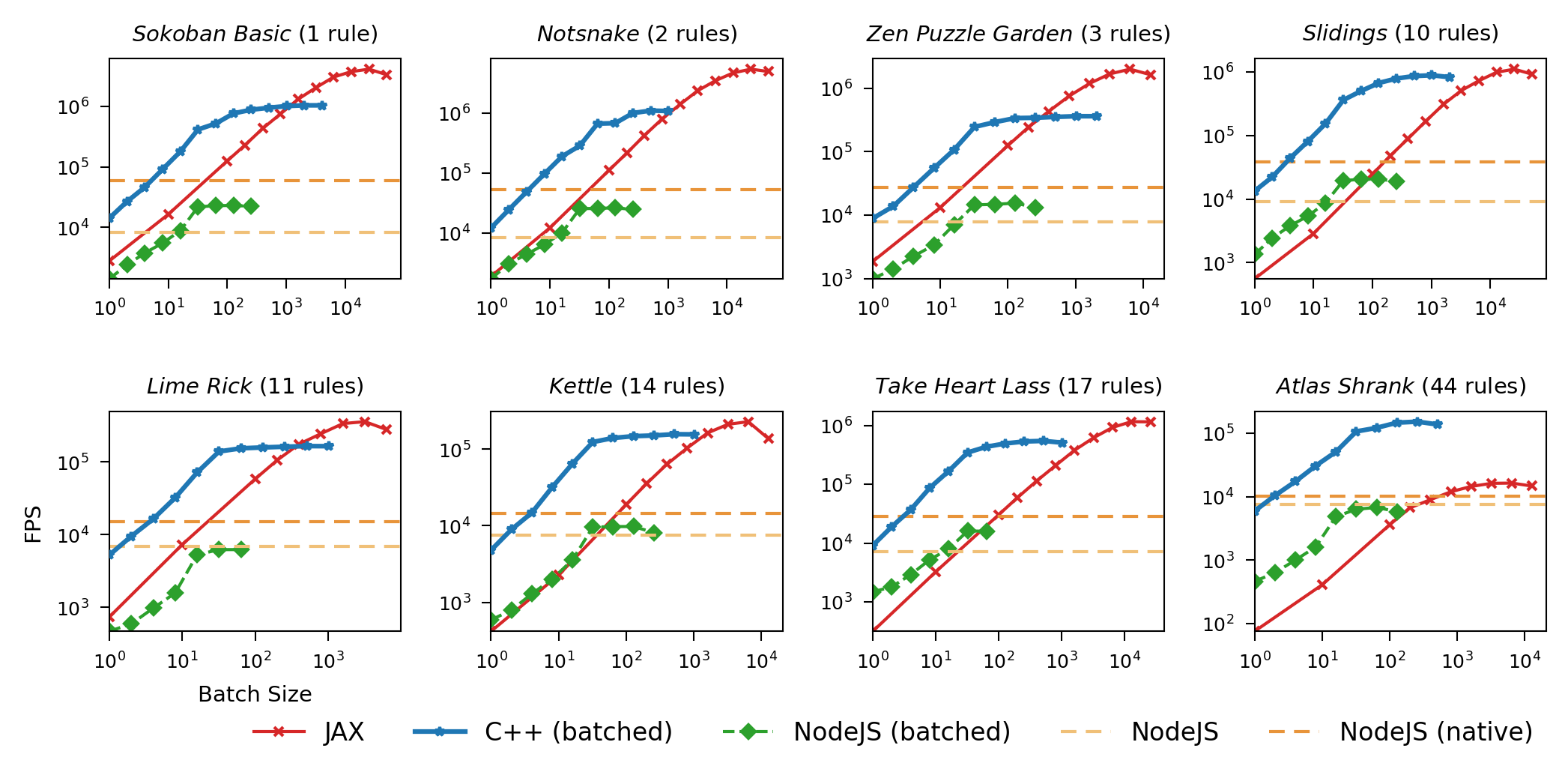}
\caption{\textbf{Parallelizing \puzzlescript for Python}. Frames-per-second of random rollouts on a 36-core i9 with 2 RTX 4090 GPUs. The original (headless) engine runs relatively fast (\textbf{NodeJS (native)}), but slows considerably when the outer loop is conducted over a Python-NodeJS bridge (\textbf{NodeJS}). Even parallelizing individual steps on the NodeJS side of the bridge (\textbf{NodeJS (batched)}) does not close the gap.
Swapping in a multi-threaded C++ reimplementation (\textbf{C++ (batched)} of the engine, however, makes for considerably faster rollouts in Python.
Running a JIT-able \textbf{JAX} reimplementation of \puzzlescript within a JIT-ed outer loop, running on the GPU, also enables significant speedups.
(For C++, we sweep over multiple \code{num\_threads} for each batch size and take the best.)
}
\label{fig:speed_profile}
\end{figure*}

%hand-waving about games as AI benchmarks

%gvgai

%griddly

%ethack (+jax version)

%crafter (+ jax version)

%that physics jax environment

%a lil history of puzzlescript

%To our knowledge, PuzzleJaxx is the first jax-compilable DSL for games.

%ScriptButler, TutoMate

\section{\puzzlescript}

\puzzlescript, published in 2013 by indie game developer increpare (Stephen Lavelle), is a description language and game engine for puzzle games. It is implemented in JavaScript and served on a public website, which includes an IDE and an interactive player interface.
% , with load and save functionality. 
% (which operates in the backend by saving and uploading games as GitHub ``gist'' files). 
Its central feature is its \emph{rewrite rules}, which describe how local patterns of objects and forces transform from one timestep to the next across the board.
. The mechanics of the classic box-pushing game Sokoban~\cite{sokoban} are defined by the following rule:\\
$[ > Player | Crate ] -> [ > Player | > Crate ]$\\
% \code{[ > Player | Crate ] -> [ > Player | > Crate ]}
% \textit{[ > Player | Crate] -> [ > Player | > Crate]}
% \lstinline[style=puzzlescript,frame=none,backgroundcolor=\color{ps-background}]{\[ > Player \| Crate \] -> \[ > Player \| > Crate \]}
% \begin{lstlisting}[style=puzzlescript,frame=none,backgroundcolor=\color{ps-background}]
% [ > Player | Crate ] -> [ > Player | > Crate ]
% \end{lstlisting}
, which indicates that whenever a Player object is in a cell next to a Crate, and moving toward the Crate, then the Crate likewise moves in this same direction.

% \puzzlescript is designed to have a quick grokking loop---once a user understands this general c

\puzzlescript games are comprised of a single file broken down into sections. The \textbf{Prelude} defines certain global parameters, like whether rules should ``tick'' at the beginning of an episode of gameplay, or whether the player must move for a tick to be considered valid.
%, or whether the play window should display the entire map or an sub-section of the map centered at the Player.

The \textbf{Objects} section defines entities---like the Player and Crate above---that may exist in the game level and interact via rewrite rules. Each object is given a name, an optional single-ASCII-character (for later use in levels), and a $5\times 5$-pixel sprite representation. 
% including a list of colors and a sprite, which is represented as a grid of integers corresponding to these colors.
The \textbf{Legend} section can be used to compositionally define meta-objects which can later be referred to in rules. For example, one might define both Player and Crate as Moveable by stating \textit{Moveable = Player or Crate}. When \textit{Moveable} appears in the left-hand-side of a rewrite rule, it indicates that \textit{either} of the component sub-objects is present in the corresponding cell. Similarly, the user can define joint-objects that can later be used to indicate the presence of \textit{both} objects simultaneously. 
% Any of these objects, or the atomic objects defined previously, can also be assigned arbitrary aliases, or ASCII shorthands.
% The \textbf{Sounds} section defines sound effects that can occur under various conditions, though we ignore it, given that sound effects in \puzzlescript games are largely auxiliary. 
% and that today's learning/reasoning paradigms generally do not leverage auditory input.
The \textbf{Collision Layers} section lists groups of objects (atomic, joint-, or meta-objects) on separate lines to indicate that these objects collide with one another and therefore cannot overlap. 
% This section can also refer to atomic objects by the above-defined short-hands or meta-objects.

The \textbf{Rules} section defines the mechanics of the game. It includes the left-right pattern rewrite rules like the ``player pushes crate'' rule above. It may also prepend these rules with keywords that define, for example, whether they only apply under certain rotations. 
% or whether they are joined together with the rules above in a single ``group'' (i.e. all applied in sequence until no rules in the group have any effect, as detailed below in the rule execution subsection).
Rule suffixes may indicate whether their application triggers a win state, a restart state (e.g. when the player walks into lava), or the repeated application of the overall tick function after the current pass.
Within rules, objects (atomic, meta- or joint-objects) may be modified by relative or absolute force indicators (``$<,>,\wedge,\vee$'' and ``left, right, up, down'' respectively) or other prefixes to indicate e.g. whether an object is stationary or absent from a given cell.
Left (input) and right (output) rule patterns may detect or project overlapping objects, respectively, though the same number of cells must be included in left and right patterns. Rules are applied in the order they are defined, and each rule is repeated until it no longer has any effect on the game state.

The \textbf{Win Conditions} section describes a set of necessary conditions which, when satisfied, result in the player ``winning'' the level. These conditions take the form: ``All ObjectA on ObjectB'', ``Some ObjectA on ObjectB'', ``No ObjectA'', or ``Some ObjectA'', indicating that all or at least one (some) of a given object must be overlapping with another object type, or that none or some of a given object type is present in the level.
Finally, the \textbf{Levels} section defines the game levels' initial layouts, using a rectangular arrangement of ASCII shorthands for objects.
This section may also define natural language messages to be displayed to the player between levels, normally used by designers to convey to the player instructions or game narrative.

% Mechanically, \puzzlescript's spatial pattern rewrite rules make it an ideal candidate for re-implementation in JAX, since they necessarily involve detecting and projecting patterns over cells of a grid, an operation which can naturally be parallelized as a series of JAX functions over fixed-size tensors.

\begin{figure*}
\centering
\includegraphics[width=\linewidth]{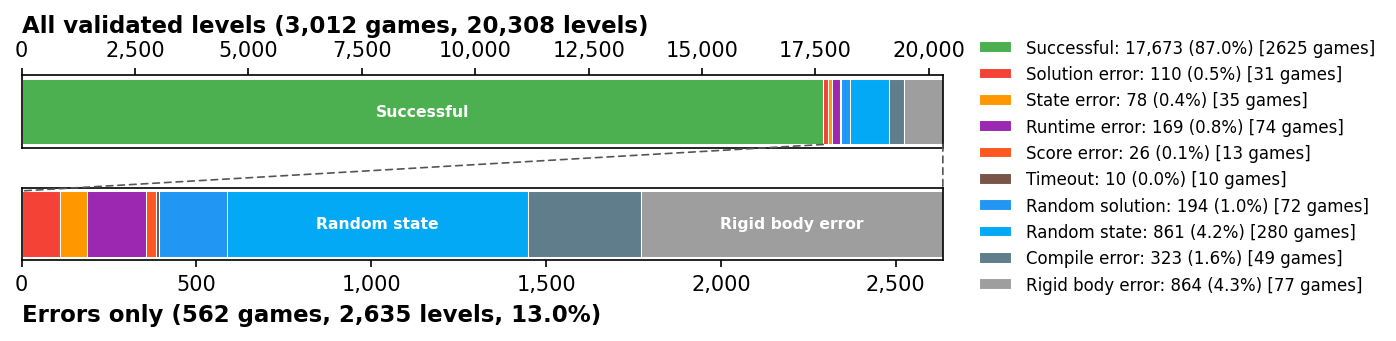}
\caption{\textbf{\name validation results}. We use search to find solutions (or best action sequences according to \textit{Win Conditions}-derived heuristics) in the original \puzzlescript engine in NodeJS, and replay action sequences in JAX, logging any discrepancies in the terminal state, its successfully triggering the win flag, terminal or intermediary score/heuristic values, and compilation errors. The majority of games in our \puzzlescript corpus are successfully validated in this fashion. The majority of solution and state errors result from randomness, and the majority of compilation errors involve the \textit{rigid} keyword, which we exclude for simplicity. (The C++ backend achieves a $>99\%$ overall success rate.)}
\label{fig:placeholder}
\end{figure*}

\section{\name Framework}
At the core of \name is the port of \puzzlescript to JAX. Our primary goal is \textit{fidelity}: to faithfully replicate the \puzzlescript engine, unifying a rich, widely-used, and challenging domain with cutting-edge advances in hardware acceleration. We focus on covering as much of \puzzlescript's feature space as possible, carefully validating implemented games and mechanics against their JavaScript counterparts to ensure identical behavior. This interoperability allows easy adoption: users and game designers can create or adapt games with the intuitive \puzzlescript web UI, then compile them into JAX learning environments without further modification. Our second goal is \textit{speed}: we aim to provide state-of-the-art throughput on a wide range of novel learning environments. \puzzlescript is a natural candidate for hardware acceleration on modern GPUs, as games are formulated entirely in terms of \textit{local rewrite rules} that modify the tile-based game state and can be applied simultaneously over the entire board. Finally, our third goal is \textit{accessibility}. We expose our new \puzzlescript backends to Python APIs, and demonstrate support for a variety of search algorithms, learning frameworks, and LLM interfaces.

\subsection{Implementing \name}
\label{sec:implement_and_validate}

First, we build a non-intrusive wrapper around a headless version of the \puzzlescript engine.
% \footnote{Any future bugfixes or changes to the JS engine can be pulled in directly.}
We implement several search algorithms---A*, (greedy) BFS, MCTS---in JS for solving arbitrary games.
Following \cite{auroriax2025PuzzleScriptPlus}, we use the DSL's Win Conditions section to infer a heuristic measuring distance from the goal state.
We can communicate with the engine from Python using a Python-NodeJS bridge, opening the possibility for integrating the \puzzlescript engine into modern AI stacks.

However, for sample-intensive and parallelizable methods like RL or evolution, the single-threaded JS engine is suboptimal.
We address this by porting PuzzleScript to JAX, hoping to take advantage of the parallel nature of the local rewrite rules at the heart of the DSL.
This us allows to jit-compile outer loops for search or optimization, i.e. a PPO player training script \cite{lu2022discovered} and an interface for search and Expert Iteration algorithms \cite{kyuseokjung2024jaxtar}.

As an alternative backend, we implement a C++ version of PuzzleScript with batched environment stepping and multi-threading to take advantage of CPU parallelism, and C++-native search algorithms.
While the JAX implementation involves substantial engineering overhead in order to reframe game logic in terms of fixed-size arrays, C++ allows for a more direct translation of the original JS code.
We provide PyTorch-based training and Expert Iteration scripts that interface with the (batched) C++ ande NodeJS engines.

The \puzzlescript DSL can be cast as a Context-Free Grammar, as in~\cite{julia2023scriptbutler}. We define a CFG in Lark~\cite{lark_github}, and use it to transform \puzzlescript game descriptions into structured Python objects for consumption by the JAX environment constructor.
Levels are represented as multihot binary arrays, with channels representing the presence of atomic objects and the directional movement or action forces that can be applied to each.

Generally, rules defined in \puzzlescript files are broken down at compile time into a \textit{Rule Group} comprising 4 rotated variants (or 2 given the rule prefixes \code{vertical} or \code{horizontal}; or 1 given the rule prefixes \code{left}, \code{right}, \code{up}, or \code{down}). The rules in a group are considered in sequence, with each applied as many times as possible (until it no longer has an effect on the level state) before moving on to the next.
Rule groups are likewise applied in sequence, each until it no longer has an effect on the level. The game file may also manually define looping rule \textit{blocks} by enclosing a sequence of rule groups in \code{startLoop} and \code{endLoop} lines, in which case the enclosed block is repeatedly executed until ineffective. Finally, a movement rule is likewise applied until it has no effect, which rule attempts to move objects one tile in the direction of any force assigned to them (removing the force if successful), attempting to apply such forces as they appear in scan-order in the level, and to objects in the order they are defined in the game's collision layers section.

This hierarchical rule execution sequence can be used to create complex dynamics between ticks of the engine, such as gravity, momentum, cellular automaton-like dynamics of growth and decay, and even multi-cell rigid body physics.
\name replicates this rule execution logic with a series of nested JAX while loops.  Wherever possible, we place logic inside python for-loops over static iterables (i.e., the number of blocks, groups within each block, and rules within each group). Though this comes at a cost in terms of compile time (as JAX effectively ``unrolls'' for-loop iterations into distinct blocks of compiled XLA code), the alternative (using a JAX \textit{switch} to select from among the list of all rule functions) significantly affects runtime speed.
We deem this to be an acceptable tradeoff given that our target is deep learning algorithms with high sample complexity.
% but the alternative---jitting these loops and using a jax \textit{switch} to select from among the list of all rule functions---effectively results in a vmapping over all these rules when the environment tick is itself batched within a vmap operation, was found to result in a significant runtime cost even given modest rulesets.
% In general, we opt for runtime over compile-time efficiency, given that our target is deep learning algorithms with high sample complexity.

\subsection{\puzzlescript Game Dataset}

We collect a dataset of over $26,000$ games saved or shared from within the \puzzlescript web editor between 2013 and 2025.
% \footnote{Originally, whenever the user saved a game in the editor, it was published to github as an anonymous ``gist''. In 2021, when GitHub began to require personal access tokens, only ``Share'' pushed to github, under the author's github username.}
(We also provide a script for continually scraping any new games shared from the editor by searching for recent GitHub gists with the automatically-generated `readme.txt'.)
We then organize the full dataset by game title and github username (or, if absent, the game's `author' field), and order them chronologically.
Interpreting these chronological commits as changes to a single game, we take the most recent of each unique game to distill the \textbf{Gists} dataset of over $8,000$ ``final'' games.

Separately, we scrape $940$ games from a community maintained \textbf{Archive} \cite{pedro2019database} (inactive since 2021) and $170$ from the current PuzzleScript \textbf{Gallery} \cite{puzzlescript_gallery}, as well as a smaller set of $15$ \textbf{Select} games (mostly from the Gallery) for select benchmarking experiments (see \autoref{fig:example_games}).
Our codebase allows running validation, benchmarking, and player agent scripts over individual games, or all of the games in any one of these datasets, allowing a natural means of scaling experiments.
In this context, we construct the datasets hierarchically, with
$$\textbf{Select} \subset \textbf{Gallery} \subset \textbf{Archive} \subset \textbf{Gists} $$
 excluding duplicate games in terms of string equivalence.

\begin{table*}[]

\begin{tabularx}{\textwidth}{|l|l|X|}
\centering
\textbf{Game} & \textbf{N. rules} & \textbf{Description} \\
\midrule
 \href{https://www.puzzlescript.net/play.html?p=748b36bc072ac4f9e173986672a942e7}{\game{Blocks}} & 0 & The player simultaneously controls multiple avatars, navigating obstacles to guide them to target positions. \\
 \hline
 \href{https://www.puzzlescript.net/editor.html}{\game{Sokoban}} &  1 & The player navigates a top-down grid of traversible and wall tiles, pushing crates onto targets. The challenge is to sequence moves such that crates do not wind up ``deadlocked'' in a position (e.g. a corner) from which they cannot be moved. \\
 \hline
\href{https://www.puzzlescript.net/editor.html}{\game{Sokoban Match 3}} & 2 & Sokoban, except that when 3 crates are arranged in a horizontal or vertical line, they disappear. \\
\hline
 \href{https://www.puzzlescript.net/editor.html}{\game{Not Snake}} & 2 & When the player moves onto a cell, they flip it between one of two colors. One must walk a hamiltonian path or strategically retrace one's steps to produce a successful grid-coloring. \\
 \hline
\href{https://www.puzzlescript.net/play.html?p=38f5c084ddc00b6262da8b8282e2122d}{\game{Zen Puzzle Garden}} & 3 & Tasks the player with ``raking'' each cell in a central square of sand without retracing its steps, while at the same time avoiding increasingly complex arrangements of obstacles within the sand patch. The player may freely navigate around the border of the sand patch. \\
 \hline
 \href{https://www.puzzlescript.net/play.html?p=6845074}{\game{Multi-Word Dictionary Game}} & 4 &  The player arranges letters by either pushing and pulling them in different directions to correctly spell an English word (``Cat'' or ``Act''). \\
 \hline
 \href{https://www.puzzlescript.net/play.html?p=9950856}{\game{Travelling Salesman}} & 4 & Player navigates a graph of nodes (projected onto the grid) with varying connectivity patterns (represented by edges connecting the border of two nodes). The player must traverse a path that touches all nodes exactly once. The player colors nodes once they traverse them, is unable to return to colored nodes, and wins once all nodes have been colored.  \\
 \hline
 \href{https://www.puzzlescript.net/play.html?p=8931824}{\game{Slidings}} (\autoref{fig:slidings}) & 10 & The player can control any one of a number of boulders (swapping between them by pressing the Action key), which they can ``slide'' in any direction until it hits an obstacle. The player must arrange these boulders onto targets in a fixed number of moves. \\
 \hline
\href{https://www.puzzlescript.net/play.html?p=6841210}{\game{Lime Rick}} (\autoref{fig:limerick} & 11 &  The player controls a caterpillar creature whose head can rise vertically by at most 3 tiles consecutively. Gravity affects the player's unsupported head and pushable crates. The player must strategically arrange their own body and pushable crates to avoid deadlock pits and climb to the exit.\\
 \hline
 \href{https://www.puzzlescript.net/play.html?p=6878681}{\game{Kettle}} (\autoref{fig:kettle}) & 14 & The player controls multiple walls of policemen, which can each move in one direction. One must strategically sequence moves to push (or ``kettle'') a group of civilians into a compact, confined square. \\
 \hline
\href{https://www.puzzlescript.net/play.html?p=6902186}{\game{Take Heart Lass}} & 17 & The player must reach the exit before it is blocked by the constantly-spreading despair. They can push pink hearts to block and delay the despair or unblock hope that spreads and consumes despair. \\
 \hline
\href{https://www.puzzlescript.net/play.html?p=6994548}{\game{Atlas Shrank}} & 44 & Platformer in which the player must reach the exit by lifting and arranging boulders to form stairs, or placing them on switches to toggle doors and platforms. \\
 \hline
\end{tabularx}
\caption{Description of the \textbf{Select} \puzzlescript games used for benchmarking.}
\label{tab:placeholder}
\end{table*}

\subsection{Experiments}

\paragraph{Speed profiling}
To compare the speed of the original \puzzlescript engine with \name, we measure frames per second under player agents taking uniformly random actions. 
All the experiments were conducted on the same consumer machine with an NVIDIA GeForce RTX 4090 GPU (for \name) and an Intel Core i9-1100K @ 3.5 GHz CPU (for \puzzlescript in NodeJS).

\paragraph{Search.} 
We perform (Greedy) Breadth-First Search, A*, and MCTS over all levels from our subset of games. We limit search to 1 million nodes.

\paragraph{Expert Iteration}
We run A* in JAXtar using the game's win heuristic $h(s)$, then train an MLP to predict
$$V(s) = \min_{s'\in T(s)}h(s')$$
 where $T(s)$ is the A*-generated sub-tree at state $s$.
We then repeat A* with this neural heuristic, minimizing
$$f(s) = g(s) + \alpha V(s) + (1 - \alpha) h(s)$$
linearly scaling $\alpha$ from $0.5$ to $1$ over 200 iterations of this 2-phase process. This is an example of Expert Iteration \cite{anthony2017thinking}, reminiscent of AlphaZero \cite{silver2018general}, in which a neural net learns to compress the information collected from prior rollouts of search into a value function.

\paragraph{RL}

Our PyTorch and (jittable) JAX (training loops randomize levels at reset.
The agent is a simple convolutional then dense feedforward network, and consumes the multihot level state as observation. In games with \verb|flickscreen| (e.g. \game{Atlas Shrank}) or \verb|zoomscreen| in the Prelude, this observation is cropped to the player-occupied subsection of the level, or an egocentric view around the player, respectively.
We compute reward as the per-timestep difference in the distance-to-win heuristic derived from the game's win conditions.
We experiment with maximum allowed episode steps of $100$ and $200$. We run 5 seeds in each setting. 

\paragraph{LLMs.}
We provide LLM player agents with an \texttt{ascii\_state} containing both the current level layout and a legend\footnote{This includes the object-to-ASCII mapping defined in the game file, in addition to any additional mappings that may be necessary as a result of the presence/overlapping of objects not seen in initial level states.}, complemented by its \texttt{rules}, alongside the agent's available \texttt{action\_space} and their \texttt{action\_meanings}. For each LLM and each game in our datasset, we perform 10 independent runs per level with a maximum episode length of $100$ steps.

\section{Results}

% \begin{figure}
% \centering
% \begin{subfigure}[t]{1\linewidth}
% \includegraphics[width=\linewidth, trim=20 0 0 20, clip]{plots/blocks_level-0.png}
% \end{subfigure}
% \begin{subfigure}[t]{1\linewidth}
% \raisebox{9 mm}{
% \includegraphics[width=.30\linewidth]{figs/blocks_rl/im-0.png}
% \includegraphics[width=.30\linewidth]{figs/blocks_rl/im-46.png}
% \includegraphics[width=.30\linewidth]{figs/blocks_rl/im-82.png}
% }
% \end{subfigure}
% \caption{In \textit{Blocks}, a PPO Reinforcement Learning agent quickly learns to increase score according to the heuristic, but falls into a sub-optimal strategy in which one of the Player blocks is trapped in a dead-end corridor adjacent to the last remaining target.}
% \label{fig:rl_curve}
% \end{figure}

\begin{figure}
\includegraphics[trim=0 4 5 5, clip, width=1.0\linewidth]{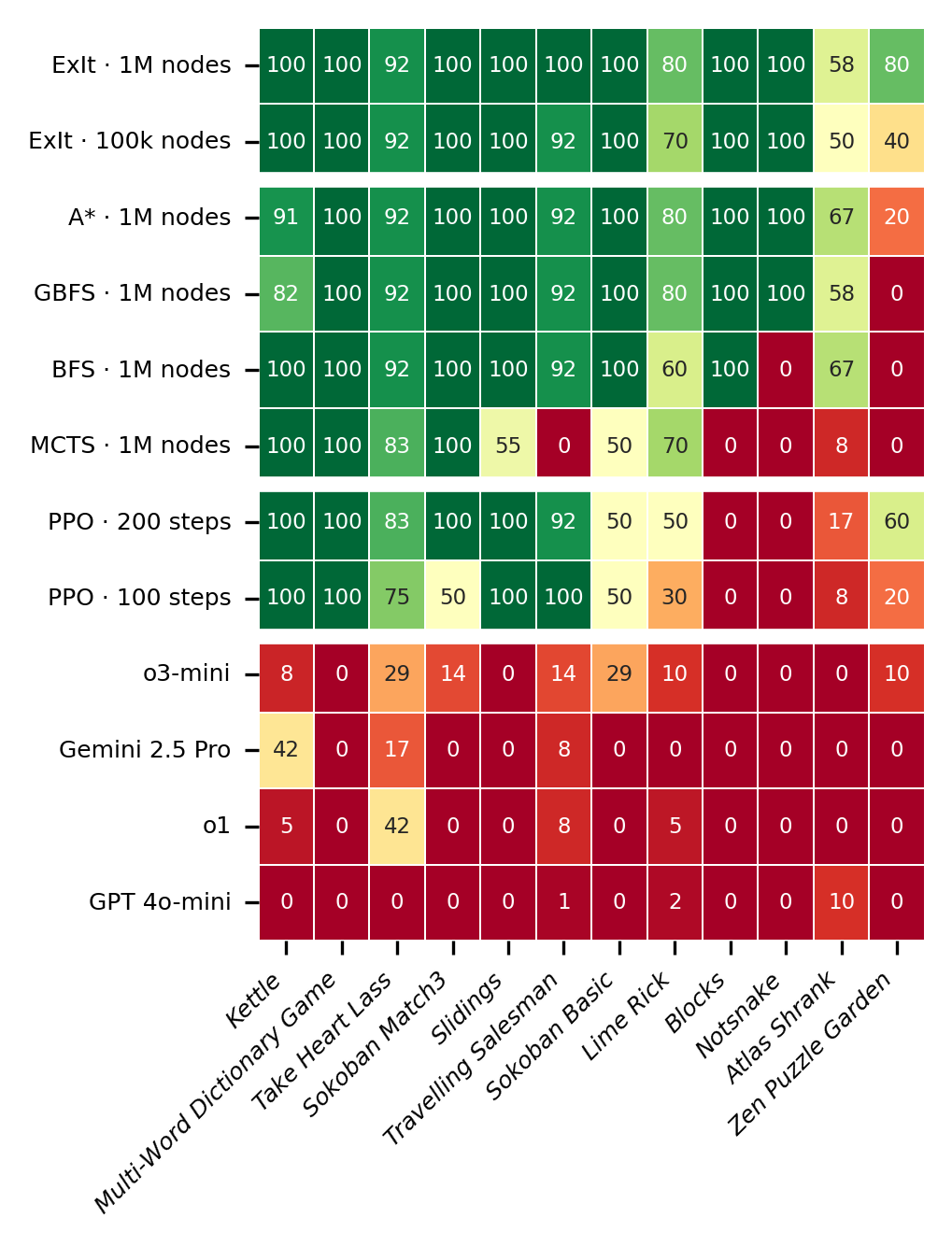}
\caption{Win rate of search, RL, and LLM-driven player agents over \name \textbf{Select}, averaged across levels. For RL (PPO) and LLMs, we additionally average across trial random seeds (5 for RL, 10 for LLMs). Games ordered by cumulative win-rate across methods. We vary the maximum nodes allotted to search and ExIt's search phase, and the maximum number of episode steps in RL.}
\label{fig:winrate_heatmaps}
\end{figure}

\paragraph{Speed profiling}
% To this end, we convert \puzzlescript into a standalone NodeJS package that can be called from Python without a browser, removing GUI-related functionality for rendering text, images, and sounds
% We profile the original engine in two settings. In one, actions are generated in Nodejs. In another, actions are generated in Python and sent to Nodejs, which better approximates the RL training scenarios targeted by \name.

In \autoref{fig:speed_profile}, we plot the number of frames per second obtained by \name on the first level of various \puzzlescript games at different batch sizes (i.e. number of environments simulated in parallel) with various backends.
Our \name backends generally achieve a $\approx10\times$ speedup over the original engine.
% The original engine is relatively fast, but is slowed considerably when actions are sent from across a Python-NodeJS bridge.
% We do not manage to make up for the difference by parallelizing multiple instances of the environment on the JS side, allowing for batched stepping from Python. A batched C++ reimplementation fares much better
JAX is superior in games with small rulesets, whereas the sequential control flow dominates in larger ones, making the multi-threaded C++ engine competitive.
The latter's lack of compile-time overhead also makes it potentially preferable for rapid iteration despite slower runtimes.

\paragraph{Tree search}

% TODO: report number of levels solved, depth of hardest levels

The poor performance of MCTS shows that random rollouts are of little utility in predicting the value of a given state in \puzzlescript. This aligns with the intuition that solutions tend to be very precise and particular, admitting little room for error or variation.
Breadth-First Search is able to completely solve half of the Select dataset, but struggles in combinatorially explosive games, like \game{Notsnake} and \game{Zen Puzzle Garden}, in which the player can ``flip'' any cell on a grid as opposed to merely displacing one of a handful of objects at a time.

% Greedy Breadth-First Search performs generally better thanks to the use of a heuristic, except in \game{Kettle}, where the heuristic would appear to be somewhat misleading. (Because the Win Condition 
% We note that the performance of tree search is very ``all-or-nothing'' as games tend to either be simple enough mechanically that brute force suffices (e.g. \textit{Sokoban} or \textit{Slidings}), or complex enough that even the simplest levels are too difficult to solve (e.g. \textit{Notsnake} or \textit{Zen Puzzle Garden}). In addition, we find that the number of search steps required in a game tends to increase as levels progress, mirroring the increasing levels of planning and problem-solving required of human players. 

% tree search on a small set of games and their constituent levels, allowing for up to 1 million environment steps or timing out after 1 minute of wall clock time, and reporting whether a solution was found, and at what iteration of search it was discovered.
% We note that many games can effectively be brute-forced by search in relatively few iterations.
% Generally, the amount of search required increases as levels progress, mirroring the increasing level of strategy and planning required by human players.

\paragraph{Reinforcement learning}

RL agents trained over the levels of a given game have partial success, but this tightly depends on dense and accurate heuristics.
\game{Atlas Shrank} is a particularly dire example.
Levels tend to be large, with solutions involving ferretting boulders to and fro to diss/reassable stair cases. But following the gradient of the Win Conditions---``\textit{All Player on Door}''---takes the player to e.g. walk down a long hallway to stand in the middle of the ground floor below a door some levels directly above, a local minimum which it is near-impossible to slip out of via random exploration.
%providing the difference between the distance-to-win heuristics derived from the game's win conditions as reward. This heuristic tries to minimize the distance between player and objects required in winning condition and between objects in the winning condition.
% Though agents quickly learn to generate increased reward, but that this learning almost always converges to incorrect solutions.
% \textit{Sokoban} and \textit{Sokoban Match 3}, while solvable via brute-force search, challenge RL agents that greedily maximize rewards but end up in deadlocked states (e.g., pushing boxes to blocked targets). In \textit{Lime Rick}, agents may lead players vertically toward the Apple but fall into pits, causing deadlocks. But these same games can be quickly brute-forced by naive breadth-first tree search.

RL performs better in games with relatively constrained state spaces and short solutions, like \game{Slidings} and \game{Kettle}, where random exploration is perhaps more likely to stumble upon correct solutions.
Increasing the maximum number of episode steps tends to improve performance, especially in \game{Zen Puzzle Garden}, where instead of producing a Hamiltonian path in one shot, the agent will often leave un-raked edge tiles then leave the garden and walk around the it to rake them at the last minute.
% ome exceptions exist, however. In \textit{NotSnake}, which has particularly dense reward (the player can increment the heuristic at each timestep by moving onto an uncolored tile), and no hard deadlocks (the player can always retrace their steps to return to a previously visited state), RL outperforms breadth-first search.
% This suggests an opportunity to improve performance on \name by combining the informed exploration of RL policies with tree search's systematic coverage of state space.

% \begin{table}[]
%     \centering
%     \input{tables/solveBFS_1000-steps}
%     \caption{Caption}
%     \label{tab:my_label}
% \end{table}

% Which games do we cover? What features are lacking? None? None!
% - [...] line detection

% A canonical set of 10, 100, and the garbage heap

% Speed profiling with random search.

% Profile with tree search.
% (With and without hashing? Expect we'll be worse with hashing, much better without. Ahmed: is there a clever way to do search with hashing on the GPU / in jax?)

% Speed profile RL. Probably it won't be worth it to try RL on vanilla PS because we've seen how slow it is above (and there is substantial overhead in converting to an RL environment accessible via python; i.e. need to rewrite PS in node.js (or use accessible-puzzlescript?)).

% Results of search.

% Results of RL training on select games.

% Results of MCTX.

\paragraph{Expert Iteration} Our hybrid search-and-learning algorithm
% , in which we cycle between performing A* with a neural heuristic and training the neural heuristic on search trajectories,
shows promise across the board. (Though more trials are needed.)
Intuitively, ExIt is a sensible direction. 
Search is useful for planning ``carefully'' in regimes like PuzzleScript where random actions are not especially meaningful and solutions tend to be exact, but heuristics in these games are often sparse (or even misleading).
The target for our neural value function arguably helps to improve the heuristic's accuracy by propagating back up the search tree from high-value leaf states.

Still, more empirical evidence would be needed before claiming that the neural heuristic currently leads to substantial improvement over a single round of A* with the environment heuristic.
Lesser performance than vanilla A* with an equal search depth in \game{Atlas Shrank} may indicate that the initial noisiness in search when factoring in the un-trained heuristic network can be detrimental.
It may also reflect an anomaly resulting from a large frontier expansion batch size in our call to JAXtar's A* implementation: updates to the set of hashed, visited states become asyncronous, and some redundant node expansions are liable to occur.

\paragraph{LLM agents}

LLMs perform meagerly across the board. (Though results need to be updated to include more recent models.)
This may well because our current RL-like framing of the task is simply a bad fit for them.
They are likely suffering for lack of a memory of prior actions, an editable persistent scratchpad and the ability to undo/restart.
Meanwhile, removing the agent's exposure to the game's source code would allow us to experiment with a more human-like interaction loop, in which agents are tasked with inferring the dynamics of the world through interaction (though it's not obvious whether this would make the task more or less challenging overall).

But at least part of the failure comes from the novelty of PuzzleScript's semantics and the raw difficulty of some of its games.
Models exhibit some success, e.g. in \game{Take Heart Lass}, where a few levels involve simply running away from spreading despair tiles and toward a goal, but get little traction on many other games.
Even \game{Multi-Word Dictionary Game} stumps these models: they presumably know how to spell ``cat'' and ``act'', but they either cannot infer the right dynamics from the game's ruleset (in which some letters are pushed and others pulled by the player), and/or plan competently around them.

\section{Discussion}

% Since these heuristics are derived automatically from a game's Win Conditions.

Puzzle games present uncommon challenges for RL and LLM-based player agents. Specifically, efficient solutions require logical inference (e.g., deduction/induction) as well as long-range planning. Even apparently simple puzzle games can be fiendishly difficult in practice. This differs qualitatively from the challenges posed by video games such as first-person shooters or platform games, where rewards are denser, and objectives and advantageous strategies may be more obvious to the player at first glance.
As such, \name, equipped with its dataset of time-tested human authored games, constitutes a unique challenge for existing game-playing AI methods.
Our empirical results---with uninformed search broadly outperforming RL and LLMs---highlight the difficulty of leveraging guided exploration and pre-trained knowledge in solving well-designed grid-based puzzles. Put differently, solving \puzzlescript games as a human would solve them, \textit{without} excessive testing of states by taking actions more or less blindly, is very much an unsolved challenge.

% ; at the same time, these are single-player games, unlike classical board games such as Chess and Go. Another main issue with puzzle games is the late rewards, where the only reward is usually if you win. This sparsity of reward might pose a challenge for RL agents. This challenge might be harder in puzzle games than in other ones with sparse rewards due to the existence of deadlock states (states where the game is still playable but not winnable after reaching them). This might pose a great challenge even for curiosity-driven agents and other techniques used to battle sparsity. 

% To avoid overfitting or over-tailoring a method to a game, it is crucial to test on a large number of games.
While testing on a large number of games is necessary to avoid overfitting in learning algorithms, humans are able to adapt to a virtually unbounded number of tasks.
To truly test the generality of artificial player agents, then, we need environment paradigms in which new, mechanically and semantically diverse tasks can be generated on demand.
\puzzlescript's DSL fills that need, and \name brings it into the modern deep learning ecosystem. 
This generative description language also opens the door to automated or partially automated design of puzzle games. This could take the form of an AI-assisted game design tool, and/or an open-ended system which combines models learning to play games with another model learning to design them, in a co-evolutionary loop.

\section{Conclusion}

% For humans, these moments can be profound. 
Our framework, \name, seeks to surface a space of simple but deep cognitive tasks for research in game-playing AI.
We reimplement \puzzlescript, an engine for grid-based puzzles, in JAX and C++ and set up a comprehensive validation loop to keep these ports in sync with the original engine.
We fuel this validation by connecting to a public datastream including thousands of games already published by its active userbase.
Additional validation scripts allow users (or coding assitants) to define relevant 
We then sample a select set of games from the public gallery and use them to benchmark various player agents.

We find that A* search can quickly solve a relatively large number of games.
Reinforcement Learning can quickly fall victim to local minima representing greedy strategies, and Large Language Models often become helplessly stuck in environments involving unconventional mechanics.
Expert Iteration, in which a neural network is repeatedly used to distill the results of A* search, then guide it, is a promising direction and shows strong initial performance.
Future attempts might combine these methods, for example using LLMs to encode additional hypothetical Win Conditions to serve as auxiliary reward.
Even our tiny set of mechanically simple Select games reveals a fine gradation of performance among different player AI approaches, and perfectly beating them all remains unattained for any approach tested thus far.

\bibliographystyle{IEEEtran}
\bibliography{refs}

@book{yannakakis2025artificial,
    title={{Artificial Intelligence and Games (Second Edition)}},
    author={Georgios N. Yannakakis and Julian Togelius},
    publisher={Springer},
    note={\url{https://gameaibook.org}},
    year={2025}
}

@incollection{szita2012reinforcement,
  title={Reinforcement learning in games},
  author={Szita, Istv{\'a}n},
  booktitle={Reinforcement Learning: State-of-the-art},
  pages={539--577},
  year={2012},
  publisher={Springer}
}

@article{todd2024gavel,
  title={GAVEL: Generating Games Via Evolution and Language Models},
  author={Todd, Graham and Padula, Alexander and Stephenson, Matthew and Piette, {\'E}ric and Soemers, Dennis JNJ and Togelius, Julian},
  journal={arXiv preprint arXiv:2407.09388},
  year={2024}
}

@article{ebner2013towards,
  title={Towards a video game description language},
  author={Ebner, Marc and Levine, John and Lucas, Simon M and Schaul, Tom and Thompson, Tommy and Togelius, Julian},
  year={2013},
  publisher={Dagstuhl Publishing}
}

@misc{auroriax2025PuzzleScriptPlus,
  title={PuzzleScriptPlus},
  author={auroriax},
  url={https://auroriax.github.io/PuzzleScript/},
  year={2025},
  note={Accessed on: March 17, 2026}
}

@misc{puzzlescript_gallery,
  title={PuzzleScript Gallery},
  author={increpare},
  url={https://www.puzzlescript.net/gallery},
  year={2015},
  note={Accessed on: March 17, 2026}
}

@misc{lark_github,
  author = {Erez Shinan},
  title = {Lark},
  year = {2025},
  url = {https://github.com/lark-parser/lark},
  note = {Accessed: March 17, 2025}
}

@article{perez2019general,
  title={General video game ai: A multitrack framework for evaluating agents, games, and content generation algorithms},
  author={Perez-Liebana, Diego and Liu, Jialin and Khalifa, Ahmed and Gaina, Raluca D and Togelius, Julian and Lucas, Simon M},
  journal={IEEE Transactions on Games},
  volume={11},
  number={3},
  pages={195--214},
  year={2019},
  publisher={IEEE}
}

@inproceedings{julia2023scriptbutler,
  title={ScriptButler serves an Empirical Study of PuzzleScript: Analyzing the Expressive Power of a Game DSL through Source Code Analysis},
  author={Julia, Clement and van Rozen, Riemer},
  booktitle={Proceedings of the 18th International Conference on the Foundations of Digital Games},
  pages={1--11},
  year={2023}
}

@misc{pedro2019database,
  author = {Pedro},
  title = {Puzzlescript games database},
  year = {2019},
  url = {https://pedros.works/puzzlescript/database},
  note = {Accessed: March 14, 2025}
}

@misc{o3-mini,
  author = {OpenAI},
  title = {OpenAI o3-mini System Card},
  year = {2025},
  month = {January},
  url = {https://openai.com/index/o3-mini-system-card/},
  note = {Accessed: March 15, 2025}
}

@MISC{sokoban,
   author = "Thinking Rabbit",
   title = "Sokoban",
   howpublished = "PC",
   year = 1982,
}

@article{bellemare2013arcade,
  title={The arcade learning environment: An evaluation platform for general agents},
  author={Bellemare, Marc G and Naddaf, Yavar and Veness, Joel and Bowling, Michael},
  journal={Journal of artificial intelligence research},
  volume={47},
  pages={253--279},
  year={2013}
}

@article{silver2016mastering,
  title={Mastering the game of Go with deep neural networks and tree search},
  author={Silver, David and Huang, Aja and Maddison, Chris J and Guez, Arthur and Sifre, Laurent and Van Den Driessche, George and Schrittwieser, Julian and Antonoglou, Ioannis and Panneershelvam, Veda and Lanctot, Marc and others},
  journal={nature},
  volume={529},
  number={7587},
  pages={484--489},
  year={2016},
  publisher={Nature Publishing Group}
}

@article{vinyals2019grandmaster,
  title={Grandmaster level in StarCraft II using multi-agent reinforcement learning},
  author={Vinyals, Oriol and Babuschkin, Igor and Czarnecki, Wojciech M and Mathieu, Micha{\"e}l and Dudzik, Andrew and Chung, Junyoung and Choi, David H and Powell, Richard and Ewalds, Timo and Georgiev, Petko and others},
  journal={nature},
  volume={575},
  number={7782},
  pages={350--354},
  year={2019},
  publisher={Nature Publishing Group}
}

@article{berner2019dota,
  title={Dota 2 with large scale deep reinforcement learning},
  author={Berner, Christopher and Brockman, Greg and Chan, Brooke and Cheung, Vicki and D{\k{e}}biak, Przemys{\l}aw and Dennison, Christy and Farhi, David and Fischer, Quirin and Hashme, Shariq and Hesse, Chris and others},
  journal={arXiv preprint arXiv:1912.06680},
  year={2019}
}

@article{duncan2011minecraft,
  title={Minecraft, beyond construction and survival},
  author={Duncan, Sean C},
  year={2011},
  publisher={Carnegie Mellon University}
}

@inproceedings{oh2016control,
  title={Control of memory, active perception, and action in minecraft},
  author={Oh, Junhyuk and Chockalingam, Valliappa and Lee, Honglak and others},
  booktitle={International conference on machine learning},
  pages={2790--2799},
  year={2016},
  organization={PMLR}
}

@article{baai2023plan4mc,
  title={Plan4mc: Skill reinforcement learning and planning for open-world minecraft tasks},
  author={BAAI, P},
  journal={arXiv preprint arXiv:2303.16563},
  year={2023}
}

@article{firoiu2017beating,
  title={Beating the world's best at Super Smash Bros. with deep reinforcement learning},
  author={Firoiu, Vlad and Whitney, William F and Tenenbaum, Joshua B},
  journal={arXiv preprint arXiv:1702.06230},
  year={2017}
}

@inproceedings{khambhayata2024mastering,
  title={Mastering Super Mario Bros.: Overcoming Implementation Challenges in Reinforcement Learning with Stable-Baselines3},
  author={Khambhayata, Manav and Singh, Shivansh and Bala, Neetu and Bhattacharyya, Arya},
  booktitle={2024 International Conference on Advances in Computing Research on Science Engineering and Technology (ACROSET)},
  pages={1--7},
  year={2024},
  organization={IEEE}
}

@article{ortega2013imitating,
  title={Imitating human playing styles in super mario bros},
  author={Ortega, Juan and Shaker, Noor and Togelius, Julian and Yannakakis, Georgios N},
  journal={Entertainment Computing},
  volume={4},
  number={2},
  pages={93--104},
  year={2013},
  publisher={Elsevier}
}

@inproceedings{torrado2018deep,
  title={Deep reinforcement learning for general video game ai},
  author={Torrado, Ruben Rodriguez and Bontrager, Philip and Togelius, Julian and Liu, Jialin and Perez-Liebana, Diego},
  booktitle={2018 IEEE conference on computational intelligence and games (CIG)},
  pages={1--8},
  year={2018},
  organization={IEEE}
}

@article{justesen2018illuminating,
  title={Illuminating generalization in deep reinforcement learning through procedural level generation},
  author={Justesen, Niels and Torrado, Ruben Rodriguez and Bontrager, Philip and Khalifa, Ahmed and Togelius, Julian and Risi, Sebastian},
  journal={arXiv preprint arXiv:1806.10729},
  year={2018}
}

@inproceedings{ojha2021cross,
  title={Cross-Game Generalization Approaches for General Video Game Playing using Deep Reinforcement Learning},
  author={Ojha, Rajesh Kumar and Rao, S Janardhana and Goel, Pankaj and Srivastava, Sandeep and Kumar, K Hareesh and Prasad, Chitturi},
  booktitle={2021 2nd International Conference on Smart Electronics and Communication (ICOSEC)},
  pages={1--8},
  year={2021},
  organization={IEEE}
}

@article{kuttler2020nethack,
  title={The nethack learning environment},
  author={K{\"u}ttler, Heinrich and Nardelli, Nantas and Miller, Alexander and Raileanu, Roberta and Selvatici, Marco and Grefenstette, Edward and Rockt{\"a}schel, Tim},
  journal={Advances in Neural Information Processing Systems},
  volume={33},
  pages={7671--7684},
  year={2020}
}

@article{hafner2021benchmarking,
  title={Benchmarking the spectrum of agent capabilities},
  author={Hafner, Danijar},
  journal={arXiv preprint arXiv:2109.06780},
  year={2021}
}

@inproceedings{matthews2024craftax,
  title={Craftax: A Lightning-Fast Benchmark for Open-Ended Reinforcement Learning},
  author={Matthews, Michael and Beukman, Michael and Ellis, Benjamin and Samvelyan, Mikayel and Jackson, Matthew Thomas and Coward, Samuel and Foerster, Jakob Nicolaus},
  booktitle={Forty-first International Conference on Machine Learning},
  year={2024}
}

@article{nikulin2024xland,
  title={XLand-minigrid: Scalable meta-reinforcement learning environments in JAX},
  author={Nikulin, Alexander and Kurenkov, Vladislav and Zisman, Ilya and Agarkov, Artem and Sinii, Viacheslav and Kolesnikov, Sergey},
  journal={Advances in Neural Information Processing Systems},
  volume={37},
  pages={43809--43835},
  year={2024}
}

@article{team2021open,
  title={Open-ended learning leads to generally capable agents},
  author={Team, Open Ended Learning and Stooke, Adam and Mahajan, Anuj and Barros, Catarina and Deck, Charlie and Bauer, Jakob and Sygnowski, Jakub and Trebacz, Maja and Jaderberg, Max and Mathieu, Michael and others},
  journal={arXiv preprint arXiv:2107.12808},
  year={2021}
}

@article{chevalier2023minigrid,
  title={Minigrid \& miniworld: Modular \& customizable reinforcement learning environments for goal-oriented tasks},
  author={Chevalier-Boisvert, Maxime and Dai, Bolun and Towers, Mark and Perez-Vicente, Rodrigo and Willems, Lucas and Lahlou, Salem and Pal, Suman and Castro, Pablo Samuel and Terry, Jordan},
  journal={Advances in Neural Information Processing Systems},
  volume={36},
  pages={73383--73394},
  year={2023}
}

@article{nasir2024gametraversalbenchmark,
  title={GameTraversalBenchmark: Evaluating Planning Abilities Of Large Language Models Through Traversing 2D Game Maps},
  author={Nasir, Muhammad Umair and James, Steven and Togelius, Julian},
  journal={arXiv preprint arXiv:2410.07765},
  year={2024}
}

@article{wu2023smartplay,
  title={Smartplay: A benchmark for llms as intelligent agents},
  author={Wu, Yue and Tang, Xuan and Mitchell, Tom M and Li, Yuanzhi},
  journal={arXiv preprint arXiv:2310.01557},
  year={2023}
}

@article{tang2025dsgbench,
  title={Dsgbench: A diverse strategic game benchmark for evaluating llm-based agents in complex decision-making environments},
  author={Tang, Wenjie and Zhou, Yuan and Xu, Erqiang and Cheng, Keyan and Li, Minne and Xiao, Liquan},
  journal={arXiv preprint arXiv:2503.06047},
  year={2025}
}

@inproceedings{paglieri2025balrog,
  title={BALROG: Benchmarking Agentic LLM and VLM Reasoning On Games},
  author={Paglieri, Davide and Cupia{\l}, Bart{\l}omiej and Coward, Samuel and Piterbarg, Ulyana and Wolczyk, Maciej and Khan, Akbir and Pignatelli, Eduardo and Kuci{\'n}ski, {\L}ukasz and Pinto, Lerrel and Fergus, Rob and others},
  booktitle={The Thirteenth International Conference on Learning Representations},
  year={2025}
}

@article{lu2022discovered,
    title={Discovered policy optimisation},
    author={Lu, Chris and Kuba, Jakub and Letcher, Alistair and Metz, Luke and Schroeder de Witt, Christian and Foerster, Jakob},
    journal={Advances in Neural Information Processing Systems},
    volume={35},
    pages={16455--16468},
    year={2022}
}

@article{Silver_2018_AlphaZero,
	title			= "A General Reinforcement Learning Algorithm that Masters Chess, Shogi, and {G}o through Self-Play",
	author			= "D. Silver and T. Hubert and J. Schrittwieser and I. Antonoglou and M. Lai and A. Guez and M. Lanctot and L. Sifre and D. Kumaran and T. Graepel and T. Lillicrap and K. Simonyan and D. Hassabis",
	journal			= "Science",
	volume			= "362",
	number			= "6419",
	pages			= "1140--1144",
	year			= "2018",
	publisher		= "American Association for the Advancement of Science"
}

@article{campbell2002deep,
  title={Deep blue},
  author={Campbell, Murray and Hoane Jr, A Joseph and Hsu, Feng-hsiung},
  journal={Artificial intelligence},
  volume={134},
  number={1-2},
  pages={57--83},
  year={2002},
  publisher={Elsevier}
}

@article{cui2025tales,
  title={TALES: Text Adventure Learning Environment Suite},
  author={Cui, Christopher Zhang and Yuan, Xingdi and Xiao, Ziang and Ammanabrolu, Prithviraj and C{\^o}t{\'e}, Marc-Alexandre},
  journal={arXiv preprint arXiv:2504.14128},
  year={2025}
}

@article{bailis2024werewolf,
  title={Werewolf arena: A case study in llm evaluation via social deduction},
  author={Bailis, Suma and Friedhoff, Jane and Chen, Feiyang},
  journal={arXiv preprint arXiv:2407.13943},
  year={2024}
}

@article{tesauro1995temporal,
  title={Temporal difference learning and TD-Gammon},
  author={Tesauro, Gerald and others},
  journal={Communications of the ACM},
  volume={38},
  number={3},
  pages={58--68},
  year={1995}
}

@article{schaeffer2007checkers,
  title={Checkers is solved},
  author={Schaeffer, Jonathan and Burch, Neil and Bjornsson, Yngvi and Kishimoto, Akihiro and Muller, Martin and Lake, Robert and Lu, Paul and Sutphen, Steve},
  journal={science},
  volume={317},
  number={5844},
  pages={1518--1522},
  year={2007},
  publisher={American Association for the Advancement of Science}
}

@inproceedings{coulom2006efficient,
  title={Efficient selectivity and backup operators in Monte-Carlo tree search},
  author={Coulom, R{\'e}mi},
  booktitle={International conference on computers and games},
  pages={72--83},
  year={2006},
  organization={Springer}
}

@article{browne2012survey,
  title={A survey of monte carlo tree search methods},
  author={Browne, Cameron B and Powley, Edward and Whitehouse, Daniel and Lucas, Simon M and Cowling, Peter I and Rohlfshagen, Philipp and Tavener, Stephen and Perez, Diego and Samothrakis, Spyridon and Colton, Simon},
  journal={IEEE Transactions on Computational Intelligence and AI in games},
  volume={4},
  number={1},
  pages={1--43},
  year={2012},
  publisher={IEEE}
}

@article{silver2010monte,
  title={Monte-Carlo planning in large POMDPs},
  author={Silver, David and Veness, Joel},
  journal={Advances in neural information processing systems},
  volume={23},
  year={2010}
}

@article{mnih2013playing,
  title={Playing atari with deep reinforcement learning},
  author={Mnih, Volodymyr and Kavukcuoglu, Koray and Silver, David and Graves, Alex and Antonoglou, Ioannis and Wierstra, Daan and Riedmiller, Martin},
  journal={arXiv preprint arXiv:1312.5602},
  year={2013}
}

@inproceedings{earle2025dreamgarden,
  title={DreamGarden: A Designer Assistant for Growing Games from a Single Prompt},
  author={Earle, Sam and Parajuli, Samyak and Banburski-Fahey, Andrzej},
  booktitle={Proceedings of the 2025 CHI Conference on Human Factors in Computing Systems},
  pages={1--19},
  year={2025}
}

@inproceedings{secretan2008picbreeder,
  title={Picbreeder: evolving pictures collaboratively online},
  author={Secretan, Jimmy and Beato, Nicholas and D Ambrosio, David B and Rodriguez, Adelein and Campbell, Adam and Stanley, Kenneth O},
  booktitle={Proceedings of the SIGCHI conference on human factors in computing systems},
  pages={1759--1768},
  year={2008}
}

@inproceedings{merino2023interactive,
  title={Interactive latent variable evolution for the generation of minecraft structures},
  author={Merino, Timothy and Charity, M and Togelius, Julian},
  booktitle={Proceedings of the 18th International Conference on the Foundations of Digital Games},
  pages={1--8},
  year={2023}
}

@inproceedings{bontrager2018deep,
  title={Deep interactive evolution},
  author={Bontrager, Philip and Lin, Wending and Togelius, Julian and Risi, Sebastian},
  booktitle={Computational Intelligence in Music, Sound, Art and Design: 7th International Conference, EvoMUSART 2018, Parma, Italy, April 4-6, 2018, Proceedings},
  pages={267--282},
  year={2018},
  organization={Springer}
}

@inproceedings{bonnet2024jumanji,
  title={Jumanji: a Diverse Suite of Scalable Reinforcement Learning Environments in JAX},
  author={Bonnet, Cl{\'e}ment and Luo, Daniel and Byrne, Donal John and Surana, Shikha and Abramowitz, Sasha and Duckworth, Paul and Coyette, Vincent and Midgley, Laurence Illing and Tegegn, Elshadai and Kalloniatis, Tristan and others},
  booktitle={The Twelfth International Conference on Learning Representations},
  year={2024}
}

@software{gymnax2022github,
  author = {Robert Tjarko Lange},
  title = {{gymnax}: A {JAX}-based Reinforcement Learning Environment Library},
  url = {http://github.com/RobertTLange/gymnax},
  version = {0.0.4},
  year = {2022},
}

@article{koyamada2023pgx,
  title={Pgx: Hardware-accelerated parallel game simulators for reinforcement learning},
  author={Koyamada, Sotetsu and Okano, Shinri and Nishimori, Soichiro and Murata, Yu and Habara, Keigo and Kita, Haruka and Ishii, Shin},
  journal={Advances in Neural Information Processing Systems},
  volume={36},
  pages={45716--45743},
  year={2023}
}

@software{kyuseokjung2024jaxtar,
  title = {JAxtar: GPU-accelerated Batched parallel A* \& Q* solver in pure JAX!},
  author = {Kyuseok Jung},
  url = {https://github.com/tinker495/JAxtar},
  year = {2024},
}

@article{anthony2017thinking,
  title={Thinking fast and slow with deep learning and tree search},
  author={Anthony, Thomas and Tian, Zheng and Barber, David},
  journal={Advances in neural information processing systems},
  volume={30},
  year={2017}
}

@article{silver2018general,
  title={A general reinforcement learning algorithm that masters chess, shogi, and Go through self-play},
  author={Silver, David and Hubert, Thomas and Schrittwieser, Julian and Antonoglou, Ioannis and Lai, Matthew and Guez, Arthur and Lanctot, Marc and Sifre, Laurent and Kumaran, Dharshan and Graepel, Thore and others},
  journal={Science},
  volume={362},
  number={6419},
  pages={1140--1144},
  year={2018},
  publisher={American Association for the Advancement of Science}
}

\appendix

\section{Use of publicly available code and data}

\name{} is based on the \puzzlescript{} game engine and Domain-Specific Language, and we further include copies of the original code within our repository for the purpose of validating existing \puzzlescript{} games in our engine.
Since \puzzlescript{} is provided with an MIT license, we include the same license in the \name{} repository. We also consulted with \name{}'s author, Stephen Lavelle, during this project's development.

To validate our engine, we used a script to scrape over 800 games from an online database~\cite{pedro2019database}, following links to Github gists containing standalone PuzzleScript game files.
We additionally validated against the games contained in the PuzzleScript Gallery\footnote{\url{https://www.puzzlescript.net/Gallery/index.html}}, and authored a wide variety of minimal test scenarios during implementation of various features.
It may also be possible to scrape games from the (currently active) PuzzleScript forum\footnote{\url{https://groups.google.com/g/puzzlescript}} (e.g. by seeking out Github gist links in threads with the ``[GAME]'' tag), or from Itch.io\footnote{\url{https://itch.io/games/made-with-puzzlescript}} (with these having the additional benefit of metadata such as user ratings, comments, and number of plays; though these do not always link to the source code in a Github gist, or do not do so in a consistent way).
Searching for PuzzleScript game file gists directly through the Github REST API may also be possible, given clever use of search keywords to circumvent pagination limits.

% In this work, we do not distribute any curated dataset of actual human-authored \puzzlescript games.
% Instead, our contribution is the \name{} engine itself.
% The set of \puzzlescript games above are used primarily to demonstrate \name{}'s coverage of a vast array of possible games, and to ensure maximum interoperability with the established \puzzlescript DSL.
% Researchers may either use the \name{} engine to run newly designed \puzzlescript-style games, or to benchmark the performance of various methods on extant \puzzlescript games, potentially drawn from one of the sources above at their own discretion.

The examplar \puzzlescript games presented in the main paper are largely drawn from the PuzzleScript Gallery, where they are presented with permission from the game authors. We list these examplar games below, with links to these games in the \puzzlescript IDE (where they are playable and editable), and authorship credits:
\begin{itemize}
\item \href{https://www.puzzlescript.net/editor.html}{Sokoban} (under Load Example → Tutorial → Basic Example) ported by Stephen Lavelle
\item \href{https://www.puzzlescript.net/editor.html}{Sokoban Match 3} (under Load Example → Tutorial → Match 3) by Stephen Lavelle
\item \href{https://www.puzzlescript.net/play.html?p=6841210}{Lime Rick} by Tommy Tuovinen
\item \href{https://www.puzzlescript.net/play.html?p=6902186}{Take Heart Lass} by Kevin Zuhn
\item \href{https://www.puzzlescript.net/play.html?p=748b36bc072ac4f9e173986672a942e7}{Blocks} by Liam K Sheehan
\item \href{https://www.puzzlescript.net/play.html?p=6878681}{Kettle} by Stephen Lavelle
\item \href{https://www.puzzlescript.net/play.html?p=6994548}{Atlas Shrank} by James Noeckel
\item \href{https://www.puzzlescript.net/play.html?p=6845074}{Multi-Word Dictionary Game} by Sarah Northway
\item \href{https://www.puzzlescript.net/play.html?p=9950856}{Travelling Salesman} by Rabbit from Hell
\item \href{https://www.puzzlescript.net/play.html?p=38f5c084ddc00b6262da8b8282e2122d}{Zen Puzzle Garden} by Lexaloffle
\item \href{https://www.puzzlescript.net/editor.html}{NotSnake} (under Load Example → Elementary → Notsnake) by Terry Cavanagh
\item \href{https://www.puzzlescript.net/play.html?p=8931824}{Slidings} by Alain Broebecker
\item \href{https://www.puzzlescript.net/editor.html}{Constellation Z} (under Load Example → Intermediate → Constellation Z) by Stephen Lavelle
\end{itemize}

\section{Limitations.}
Though most of the major features of \puzzlescript are replicated in \name, we identify in our dataset of human games some edge cases which are fail validation against the NodeJS version of the original engine.
This is true of all games involving randomness, since random seeds cannot be controlled and aligned between NodeJS and JAX.
% There are also idiosyncracies in the way game states in games with over 32 unique objects are represented in NodeJS, which causes inconsistencies when comparing against game states in JAX; though this can be easily resolved in theory.
Some games surface issues in our implementation which still need to be addressed, for example by violating our definition of the \puzzlescript DSL as a context-free grammar or causing compile or runtime issues in our JAX environment.
At the same time, having been designed with fidelity as a first priority, further speed optimizations are almost certainly possible. 
% for example by pre-co
Meanwhile, we apply only simple, off-the-shelf algorithms to our domain in this preliminary study.
More sophisticated RL algorithms with more robust exploration strategies, or more comprehensive LLM prompting strategies including relevant history of prior game states, could likely be used to improve performance.

\section{Ethical considerations}

\name{} is intended as a benchmark to assist in developing more generally capable and human-like AI agents, in particular by surfacing questions about the role of insight to solve a mechanically and semantically rich space of diverse puzzle games.
We acknowledge that the overarching goal of creating generally capable AI agents may present both dangers and benefits to humanity.
While these broader questions are out of scope for the present discussion, we believe that benchmarks like \name{} are crucial in understanding AI agents and learning algorithms which appear to have super-human abilities in some domains, but whose limitations are often poorly understood.
\name{} is particularly relevant because it brings to the fore a swath of domains in which we expect many state-of-the-art agents and algorithms are likely to fail in surprising and perhaps counter-intuitive ways, even despite the apparent simplicity of the tasks at hand.

\puzzlescript's DSL makes it easy, for example, to invert the canonical semantics of a game like \textit{Sokoban}, such that with a simple variation to the game's rules, the player now pushes a crate forward by moving \textit{away} from it (as in \href{https://www.puzzlescript.net/play.html?p=3fc507f08c079b1da607c2d6dc7ee8a4}{Okosban}).
We expect that in games with such inverted or otherwise alien semantics, LLMs may have particularly difficulty in generating competent strategies (even supposing a more robust LLM-player pipeline is developed to address their difficulties in solving more canonical puzzles).
As such \name{} can serve as an effective test of the abilities of LLMs to reason and problem-solve in the kind of out-of-distribution scenarios they may encounter once deployed into the wild, which situations may ultimately be of high consequence of users and designers.

In terms of \name{}'s impact on game designers, we hope that by fostering the development of more capable puzzle-solving agents, designers of \puzzlescript{} games may eventually be able to automatically playtest their games more effectively.
\puzzlescript's creator has recently expressed apprehension around embedding a best-first-search-driven solver agent~\footnote{Available at \url{https://github.com/Auroriax/PuzzleScriptPlus/blob/master/README.md}.} into the \puzzlescript{} IDE, given that it might lead designers to create games that are significantly complex or challenging from the perspective of tree search, but potentially un-interesting or less fun or enjoyable for human players\footnote{\url{https://x.com/increpare/status/1905568607410532690}}.
Given that \name{} facilitates the development of a wide variety of AI player agents beyond simple tree search---such as those based on LLMs, or those involving Reinforcement Learning---we hope that developers might ultimately have access to a diverse set of potentially human-like agents, allowing them to automatically measure proxies of human enjoyment or satisfaction (granted, this will likely require significant algorithmic advances, and benchmarking any such proxy benchmarks against actual human playtraces and surveys).

As alluded to in our Conclusion, \name{} also potentially facilitates the use of LLMs or genetic programming to generate new puzzle games automatically (e.g. by leveraging metrics generated by diverse player agents inside an evolutionary loop, as in~\cite{todd2024gavel}).
Concerns may be raised here around the potential for automating away the process of game design, and burying human ingenuity and artistry in a barrage of AI-generated content that maximizes superficial metrics of player retention or engagement.
In this regard, we advocate for the development of design assistant tools that incorporate human feedback and allow designers to intervene in the process of automatic game generation, as in \cite{earle2025dreamgarden}, or as in the general paradigm of design through interactive evolution~\cite{secretan2008picbreeder,merino2023interactive,bontrager2018deep}.

\section{Additional implementation and validation details}

% \begin{table}
% \caption{Results of validating \puzzlescript games in \name, by using breadth-first search to generate solutions for each level in JavaScript, then replaying these solutions in JAX, and ensuring they lead to equivalent end-states.}
% \centering
% \input{tables/validation_results}
% \vspace{0.3cm}
% \label{tab:validation_results}
% \end{table}

To validate the fidelity of \name{}, we use breadth-first search to find solutions for each level of each game in our collected dataset.
We cap the number of environment steps during search at 100,000 and set a timeout of 1 minute.
Where search does not find a winning state, we return the action sequence leading to the highest score, and in case of ties prefer longer action sequences (in hopes of exploring more of the game's state space and thus ensuring a more robust validation).
% (The full results of this search procedure on the collected dataset of games is reported in \autoref{tab:tree_search}.)
We then initialize each game and level in \name, and replay the action sequence, ensuring that it results both in the win conditions being met, and in an equivalent state (in terms of the layout of object in the level).

The one major feature which, to our knowledge, remains unimplemented in \name{} is the ``rigid'' keyword, which is used to simulate rigid-body physics.
The use of this keyword appears in only $9$ games in our dataset (leading to $9$ compilation errors).
We omit it for simplicity, given that its implementation appears relatively involved, and would require the use of additional channels in our level-state representation.
The \puzzlescript documentation stresses this point, in fact (with the author writing that they ``kinda regret adding this keyword to the engine'') and strongly advises the user to deploy other strategies to simulate rigid-body physics\footnote{\url{https://www.puzzlescript.net/Documentation/rigidbodies.html}}.

In addition to the features described in the body of this paper, we note that we also implement the ``line detector'' feature (denoted in the \puzzlescript DSL as an ellipsis), which recognizes patterns separated by an arbitrary number of tiles along a row or column.
Under the hood, we treat line detectors as a special kind of kernel that detect sub-kernels (the groups of cells on either side of the ellipsis) across the board, then detect if these subkernels' activations fall in some ordered sequence along a line.
These sequences are considered in order of the least to the most space between the subkernels.
The line projection function then iterates through these detected lines in order, attempting to apply their respective subkernels until this has an effect on the board.

\section{Additional Results}

\begin{figure*}
    \centering
    \includegraphics[width=1.0\linewidth,trim=0 20 120 0,clip]{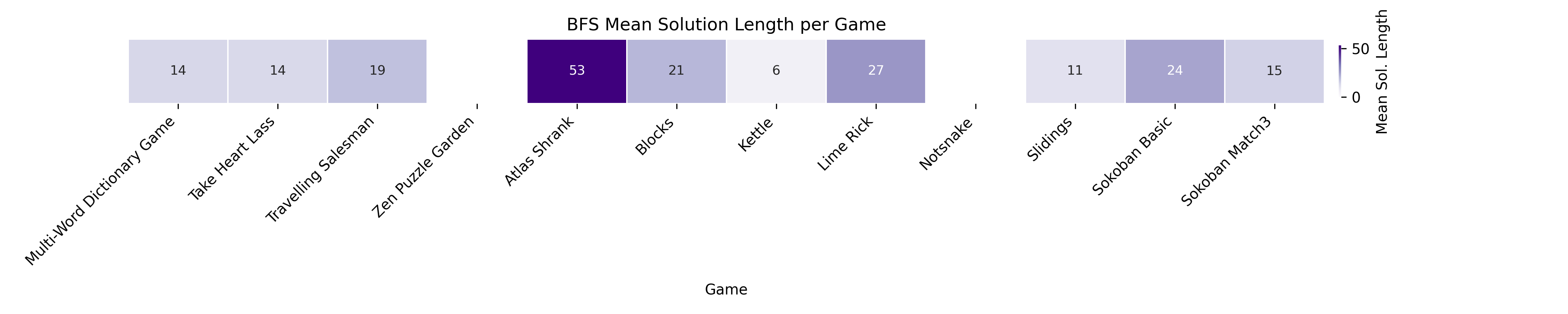}
    \caption{Average length of solutions generated by BFS over our test suite of \name games.}
    \label{fig:bfs_sol_length}
\end{figure*}

\begin{figure}
\begin{subfigure}{\linewidth}
    \centering
    \includegraphics[width=1.0\textwidth,trim=0 0 0 0,clip]{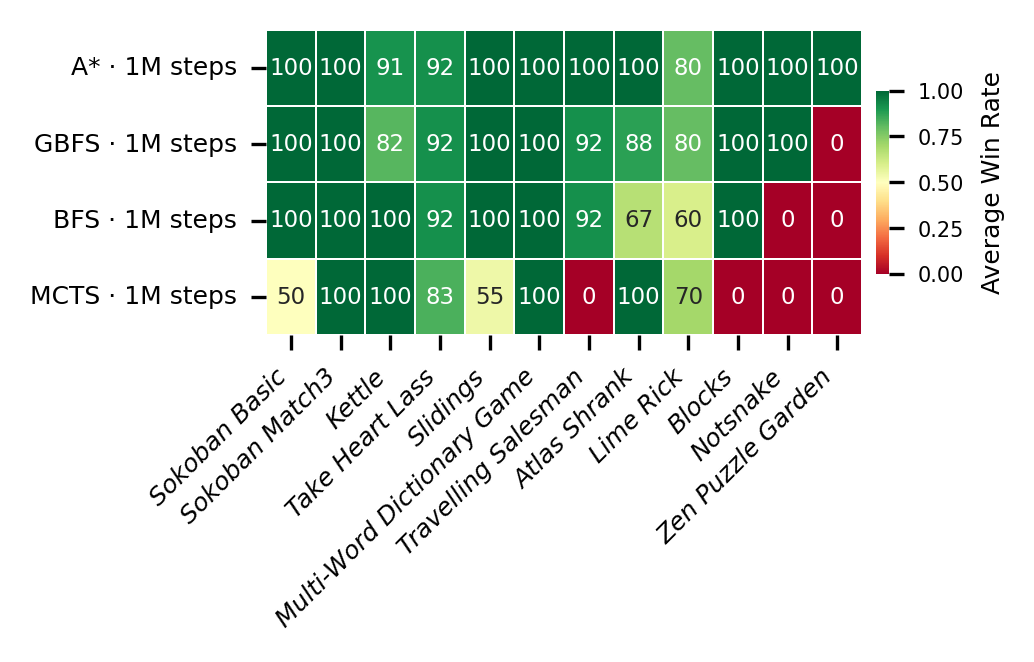}
    \caption{Average win rates of search.}
    \label{fig:bfs_heatmap}
\end{subfigure}
\begin{subfigure}{\linewidth}
    \centering
    \includegraphics[width=1.0\linewidth,trim=0 20 120 0,clip]{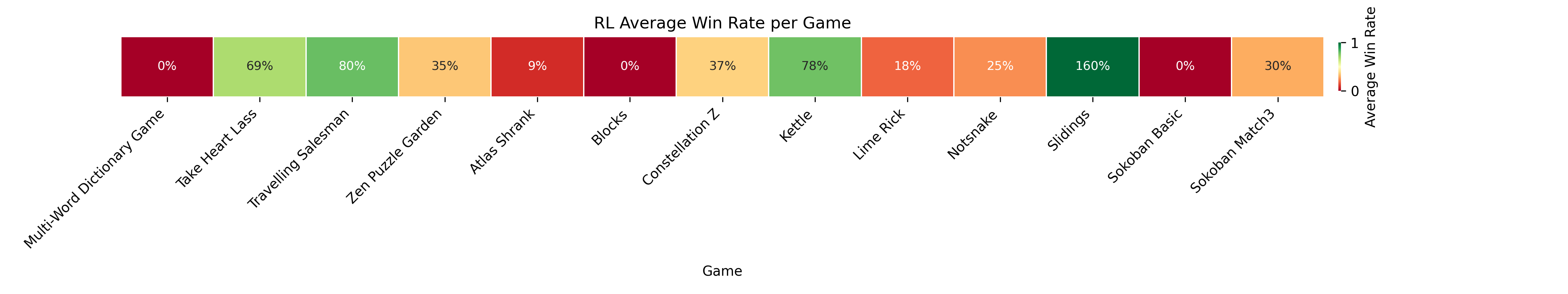}
    \caption{Average win rates of RL (PPO).}
    \label{fig:bfs_heatmap}
\end{subfigure}
\begin{subfigure}{\linewidth}
\centering
\includegraphics[width=1\textwidth, trim=22 32 210 32, clip]{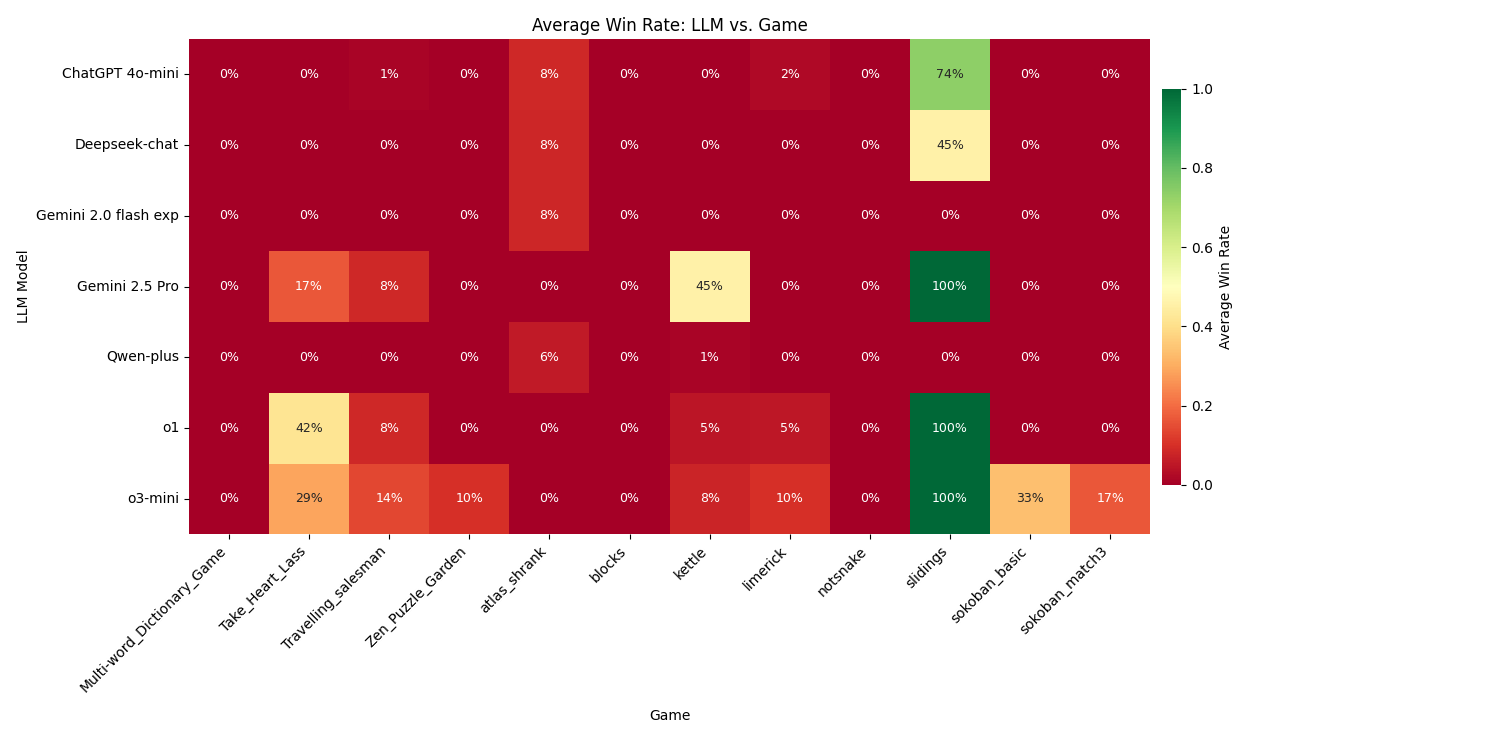}
\caption{Average win rates of LLM-driven player agents.}
\label{fig:heatmap_results}
\end{subfigure}
\caption{Win rate of search, RL, and LLM-driven player agents over a subset of \name games, averaged across levels.}
\label{fig:winrate_heatmaps}
\end{figure}

\subsection{LLMs}

For our LLM experiments, we employed both reasoning-enabled LLMs and non-reasoning LLMs. Based on the experimental results presented in \autoref{fig:heatmap_results}, we observe significant performance variations across different LLMs when evaluated on 12 distinct games compiled in \name. The findings reveal that all model performance is highly task-dependent, with no single model demonstrating consistent superiority across all evaluated games. Notably, \texttt{o3-mini} achieved perfect performance (100\% win rate) on the \textit{Slidings} puzzle task and demonstrated strong capabilities in several other games, including \textit{Sokoban Basic} (50\%), \textit{Take Heart Lass} (23\%), \textit{Travelling Salesman}, and \textit{Sokoban match 3} (25\%). \texttt{DeepSeek-R1} exhibited exceptional performance on the \textit{Slidings} puzzle task (91\% win rate) while showing moderate success in strategic games such as \textit{Take Heart Lass} (25\%) and \textit{Travelling salesman} (8\%). \texttt{ChatGPT-4o-mini} displayed a more balanced performance profile, achieving its highest success rate on the \textit{Slidings} puzzle (74\%) and moderate performance on \textit{Atlas Shrank} (8\%) and \textit{Limerick} (2\%). In contrast, models such as \texttt{Qwen-plus} and \texttt{Gemini} showed limited success across most tasks, with \texttt{Qwen-plus} achieving only 6\% on \textit{Atlas Shrank} and 1\% on \textit{Kettle}, while \texttt{Gemini}'s performance peaked at 8\% on \textit{Atlas Shrank}. The results suggest that certain games, particularly \textit{Slidings} puzzles, may be more amenable to current language model capabilities, while others such as \textit{Multi-Word Dictionary Game}, \textit{Blocks}, \textit{Notsnake}, and \textit{Zen Puzzle Garden}, remain challenging across all evaluated models.

\subsection{Human Playtesting}
\label{sec:human}

\begin{figure}
\begin{subfigure}[t]{0.49\linewidth}
    \centering
    \includegraphics[width=1.0\linewidth]{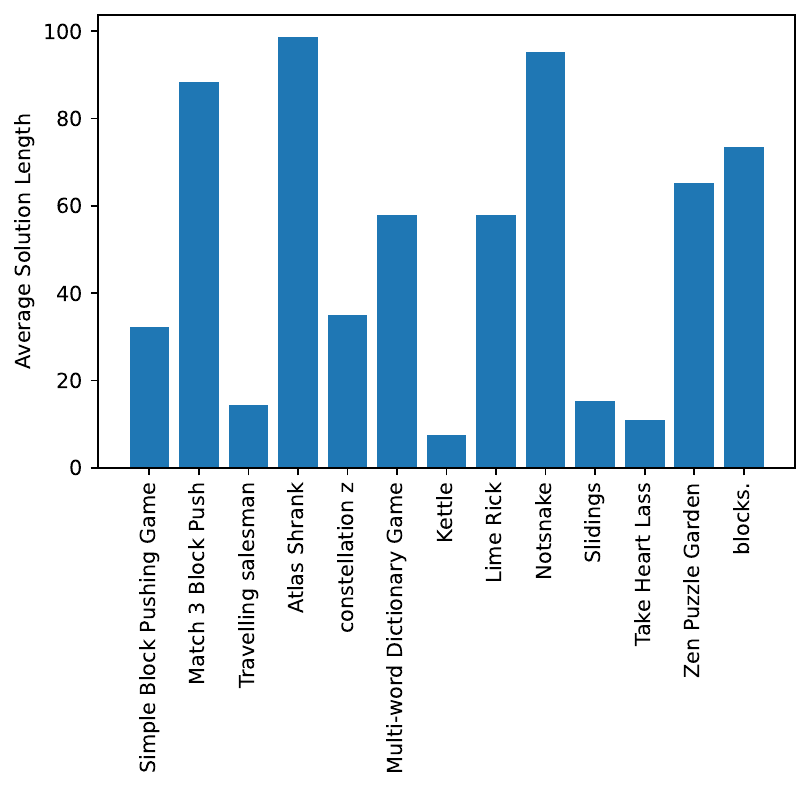}
    \caption{Average solution length (number of actions) in solutions generated by human players over our suite of \puzzlescript games (averaged over users, and levels for each game).}
    \label{fig:human_sol_length}
\end{subfigure}\hfill
\begin{subfigure}[t]{0.49\linewidth}
    \includegraphics[width=1.0\linewidth]{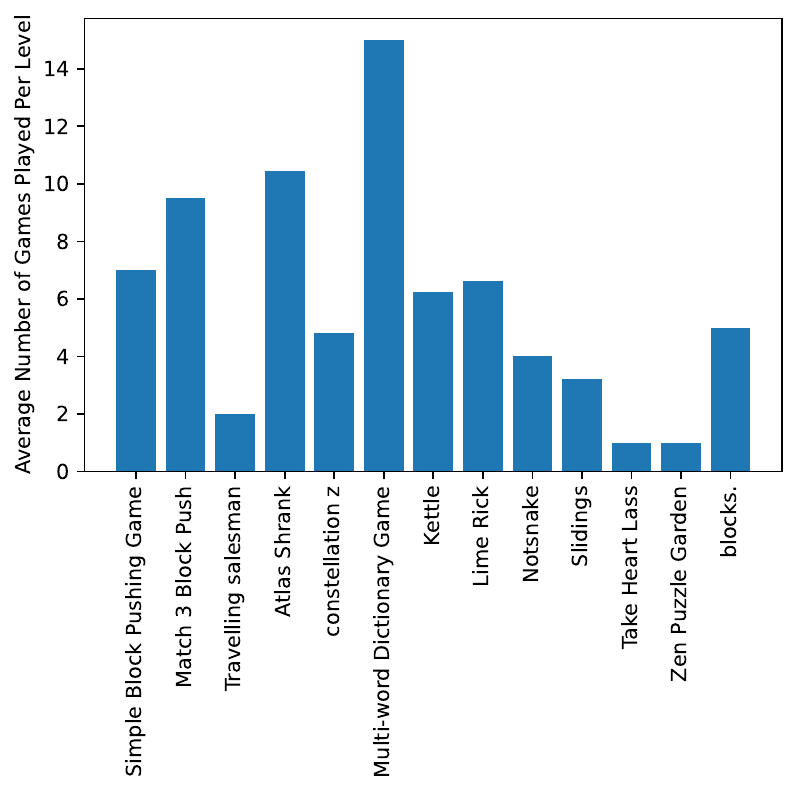}
    \caption{Average number of play-throughs for each game, summed over users and averaged across levels.}
    \label{fig:human_playthroughs}
\end{subfigure}
\end{figure}

To gain a sense of the tractability of \puzzlescript games for human players, we run a preliminary online user study in which users are invited to play one or any of the games from our test suite using a fork of the \puzzlescript web player, modified to send successful solutions to a database.
We find that overall, the trends in terms of the differences in solution lengths between games (\autoref{fig:human_sol_length}) mirror those exhibited in the optimal solutions returned by breadth-first search (\autoref{fig:bfs_sol_length}).

\begin{table}[]
    \caption{Results from preliminary playtesting of \puzzlescript games with human players.}
    \label{tab:placeholder}
    \centering
    \begin{tabular}{lrrr}
\hline
 Game                       &   \# Levels &   Total Plays &   Avg Solution Length \\
\hline
 Simple Block Pushing Game  &          2 &            14 &                 32.36 \\
 Match 3 Block Push         &          2 &            19 &                 88.47 \\
 Travelling salesman        &          6 &            12 &                 14.33 \\
 Atlas Shrank               &          9 &            94 &                 98.79 \\
 constellation z            &         11 &            53 &                 35.02 \\
 Multi-word Dictionary Game &          1 &            15 &                 57.93 \\
 Kettle                     &          8 &            50 &                  7.52 \\
 Lime Rick                  &         11 &            73 &                 57.82 \\
 Notsnake                   &          1 &             4 &                 95.25 \\
 Slidings                   &          5 &            16 &                 15.38 \\
 Take Heart Lass            &          5 &             5 &                 11    \\
 Zen Puzzle Garden          &          5 &             5 &                 65.2  \\
 blocks.                    &          1 &             5 &                 73.6  \\
\hline
\end{tabular}
\end{table}

\subsection{Reinforcement learning}

For our Reinforcement Learning experiments, we use the fully-jitted training loop written in JAX provided by \cite{lu2022discovered}, allowing us to take advantage of \name's jitted environment step function.
(We add utilities for saving model checkpoints and rendering episodes intermittently during training.)
We use the above repo's default hyperparameters for PPO, training agents on each level over $5$ different random seeds for a total of $5$ million environment steps each, with a learning rate of $1e^{-4}$, $128$ rollout steps per minibatch, with $4$ minibatches and $10$ update epochs, with a $\gamma=0.99$, an entropy coefficient of $0.01$ and a value function coefficient of $0.5$.
We set batch size as large as possible for each game and level combination within the constraints of the VRAM available on the GPUs we use for training.

We use our institution's high-performance computing cluster for training, and include in our codebase scripts for deploying sweeps of training jobs to nodes in this cluster via SLURM (we provide similar scripts in order to parallelize the tree search and JAX episode-rollouts in our \name validation pipeline).
The GPUs on this cluster include the NVIDIA RTX8000, V100, A100, and H100, and the AMD MI100 and MI250.
(We use a separate consumer machine with an NVIDIA 4090 for our speed profiling experiments).

\begin{figure}
\centering
\begin{subfigure}[t]{0.49\linewidth}
\includegraphics[width=\linewidth]{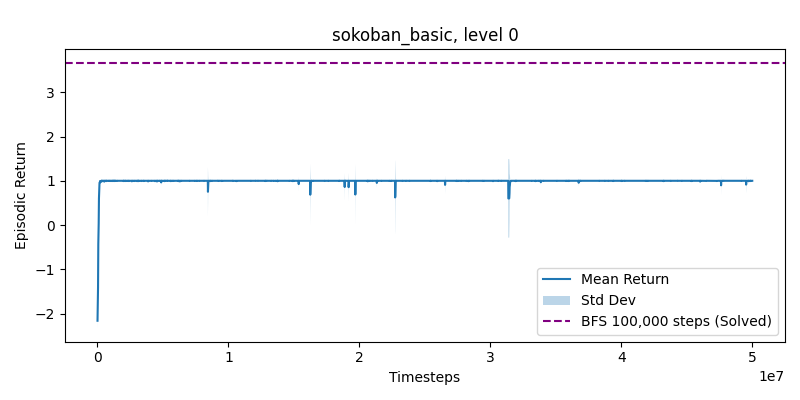}
\end{subfigure}
\begin{subfigure}[t]{0.49\linewidth}
\raisebox{9 mm}{
\includegraphics[width=0.3\linewidth]{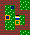}
\includegraphics[width=0.3\linewidth]{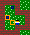}
\includegraphics[width=0.3\linewidth]{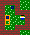}
}
\end{subfigure}
\begin{subfigure}[t]{0.49\linewidth}
\includegraphics[width=\linewidth]{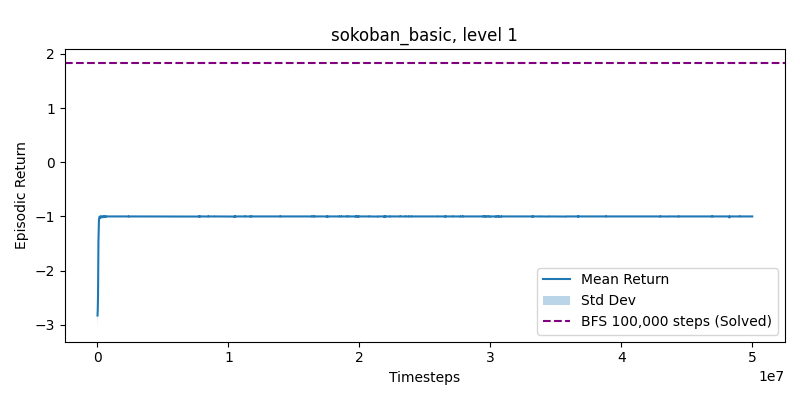}
\end{subfigure}
\begin{subfigure}[t]{0.49\linewidth}
\raisebox{9 mm}{
\includegraphics[width=0.3\linewidth]{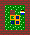}
\includegraphics[width=0.3\linewidth]{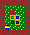}
\includegraphics[width=0.3\linewidth]{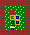}
}
\end{subfigure}
\caption{Comparison of RL against breadth-first search in \textit{Sokoban}.
Episode rollouts from RL are pictured on the right. Here, the agent greedily maximizes the heuristic (the sum of manhattan distances between targets and their nearest crates), preventing discovery of optimal solutions.
}
\label{fig:rl_sokoban}
\end{figure}

\begin{figure}[t]
\centering
\begin{subfigure}[t]{0.49\linewidth}
\includegraphics[width=\linewidth]{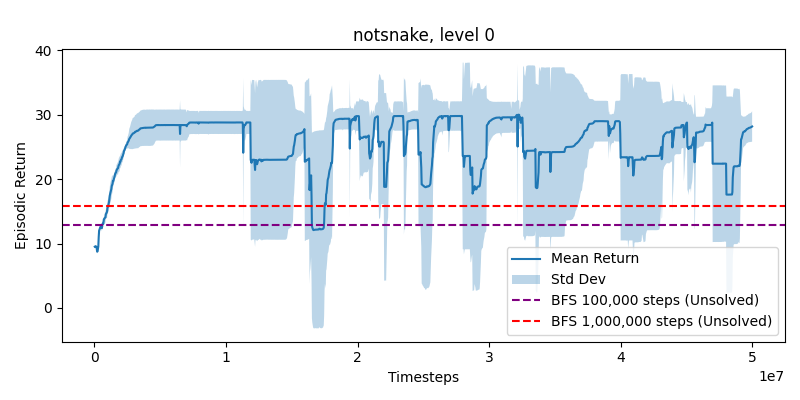}
\end{subfigure}
\begin{subfigure}[t]{0.49\linewidth}
\raisebox{9 mm}{
\includegraphics[width=0.3\linewidth]{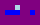}
\includegraphics[width=0.3\linewidth]{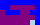}
\includegraphics[width=0.3\linewidth]{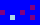}
}
\end{subfigure}
\caption{Comparison of RL against breadth-first search in \textit{Notsnake}.
Episode rollouts from RL is pictured on the right.
}
\label{fig:rl_notsnake}
\end{figure}

\begin{figure}
\centering
\begin{subfigure}[t]{0.49\linewidth}
\includegraphics[width=\linewidth]{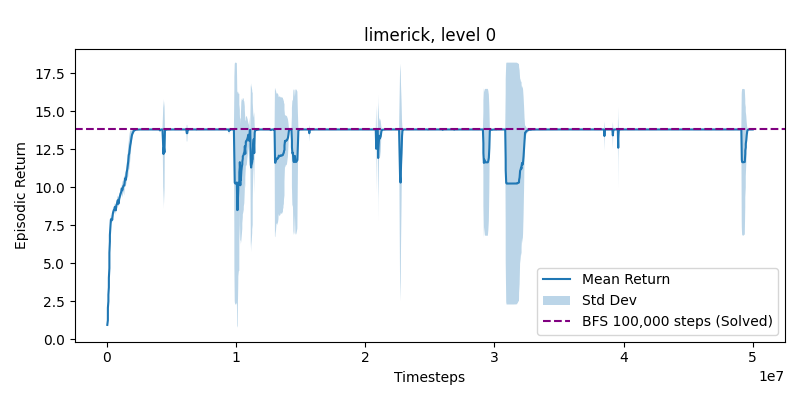}
\end{subfigure}
\begin{subfigure}[t]{0.49\linewidth}
\raisebox{9 mm}{
\includegraphics[width=0.3\linewidth]{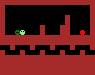}
\includegraphics[width=0.3\linewidth]{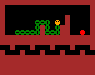}
\includegraphics[width=0.3\linewidth]{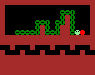}
}
\end{subfigure}
\begin{subfigure}[t]{0.49\linewidth}
\includegraphics[width=\linewidth]{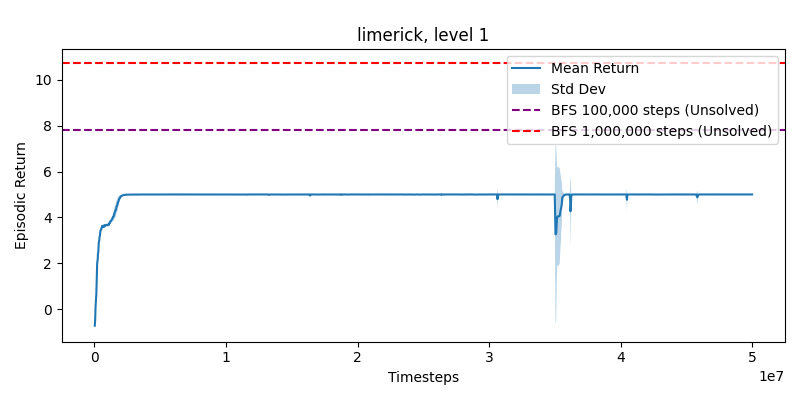}
\end{subfigure}
\begin{subfigure}[t]{0.49\linewidth}
\raisebox{9 mm}{
\includegraphics[width=0.3\linewidth]{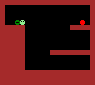}
\includegraphics[width=0.3\linewidth]{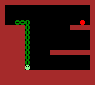}
\includegraphics[width=0.3\linewidth]{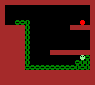}
}
\end{subfigure}
\begin{subfigure}[t]{0.49\linewidth}
\includegraphics[width=\linewidth]{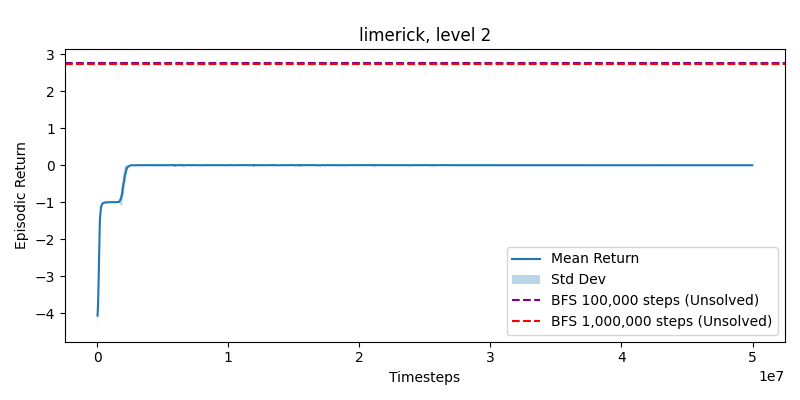}
\end{subfigure}
\begin{subfigure}[t]{0.49\linewidth}
\raisebox{9 mm}{
\centering
\includegraphics[width=0.3\linewidth]{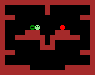}
% \hspace{1cm}
\includegraphics[width=0.3\linewidth]{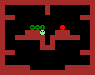}
}
\end{subfigure}
\begin{subfigure}[t]{0.49\linewidth}
\includegraphics[width=\linewidth]{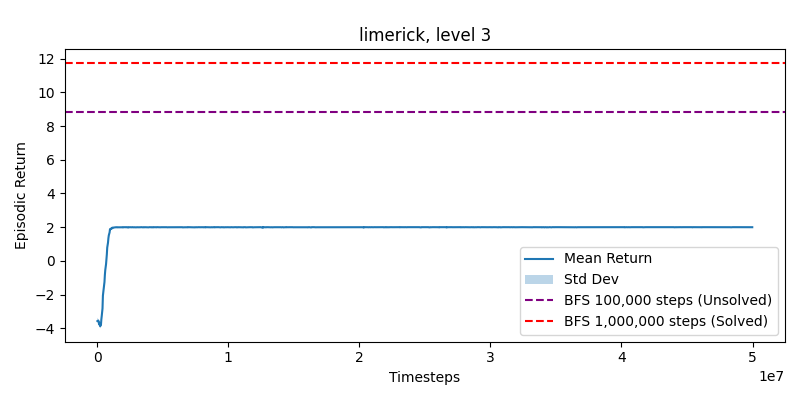}
\end{subfigure}
\begin{subfigure}[t]{0.49\linewidth}
\raisebox{9 mm}{
\includegraphics[width=0.3\linewidth]{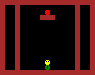}
\includegraphics[width=0.3\linewidth]{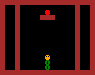}
\includegraphics[width=0.3\linewidth]{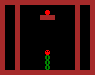}
}
\end{subfigure}
\begin{subfigure}[t]{0.49\linewidth}
\includegraphics[width=\linewidth]{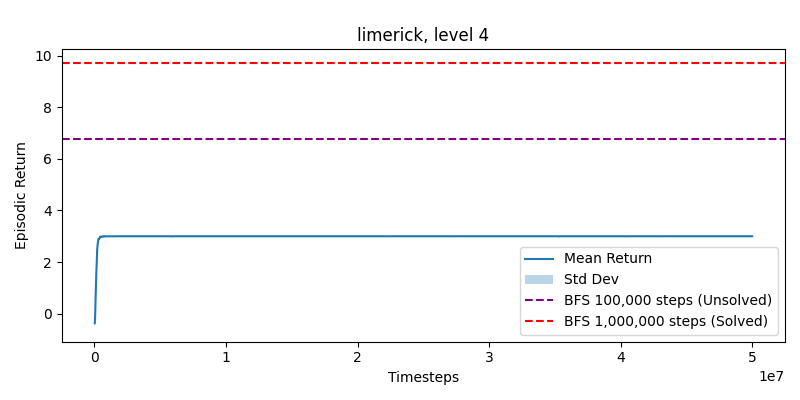}
\end{subfigure}
\begin{subfigure}[t]{0.49\linewidth}
\raisebox{9 mm}{
\includegraphics[width=0.3\linewidth]{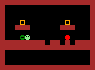}
\includegraphics[width=0.3\linewidth]{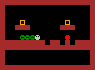}
\includegraphics[width=0.3\linewidth]{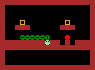}
}
\end{subfigure}
\begin{subfigure}[t]{0.49\linewidth}
\includegraphics[width=\linewidth]{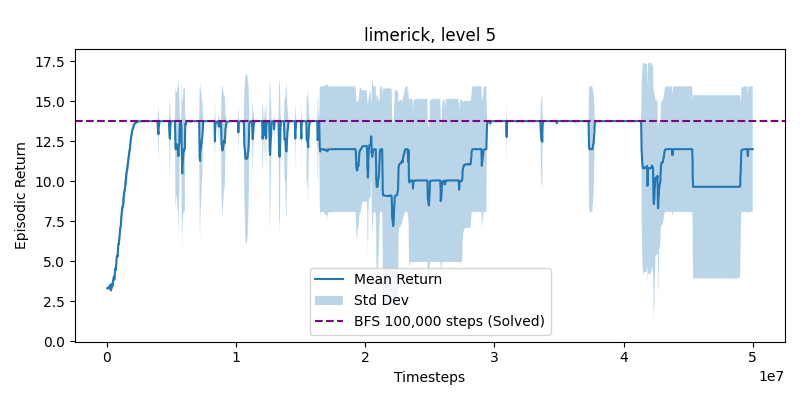}
\end{subfigure}
\begin{subfigure}[t]{0.49\linewidth}
\raisebox{9 mm}{
\includegraphics[width=0.3\linewidth]{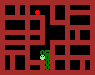}
\includegraphics[width=0.3\linewidth]{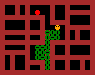}
\includegraphics[width=0.3\linewidth]{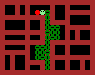}
}
\end{subfigure}
\caption{Comparison of RL against breadth-first search in \textit{Limerick}. Episode rollouts from RL are pictured on the right. Agents only master levels with a relatively straightforward path to the goal. They do not generally uncover strategies involving significant roundabouts away from the goal, and can fall prey to ``obvious'' traps along the more direct path.}
\label{fig:rl_limerick}
\end{figure}

\begin{figure}
\centering
\begin{subfigure}[t]{0.49\linewidth}
\includegraphics[width=\linewidth]{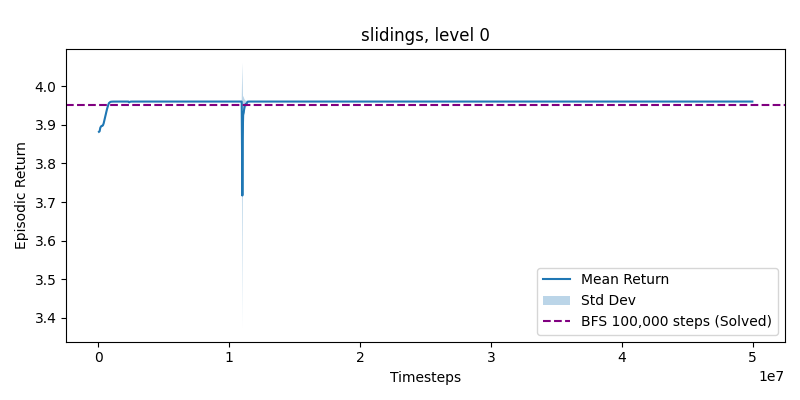}
\end{subfigure}
\begin{subfigure}[t]{0.49\linewidth}
\includegraphics[width=\linewidth]{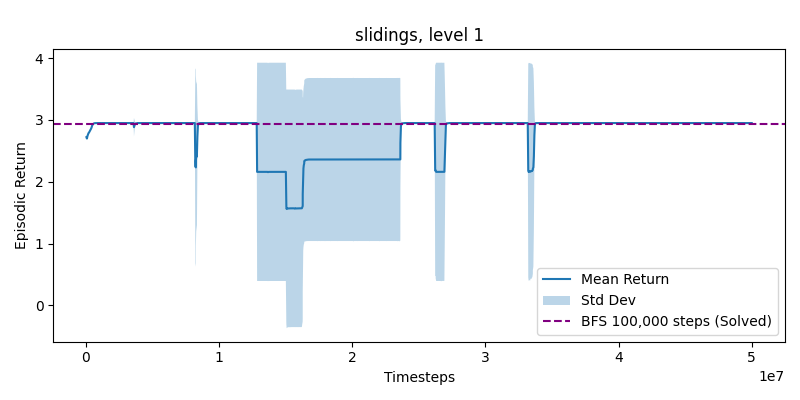}
\end{subfigure}
\begin{subfigure}[t]{0.49\linewidth}
\includegraphics[width=\linewidth]{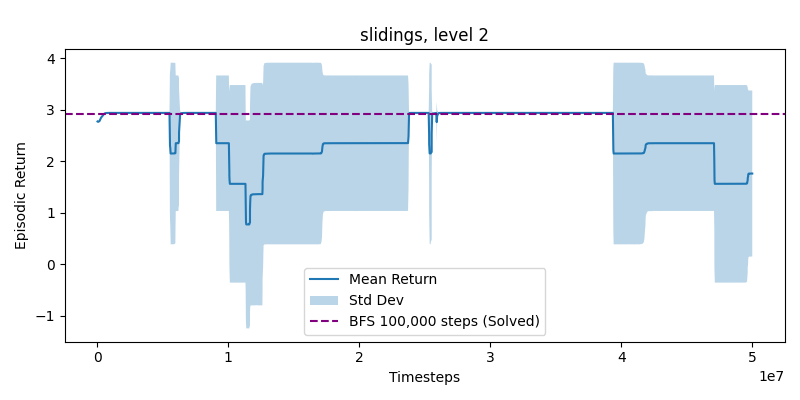}
\end{subfigure}
\begin{subfigure}[t]{0.49\linewidth}
\includegraphics[width=\linewidth]{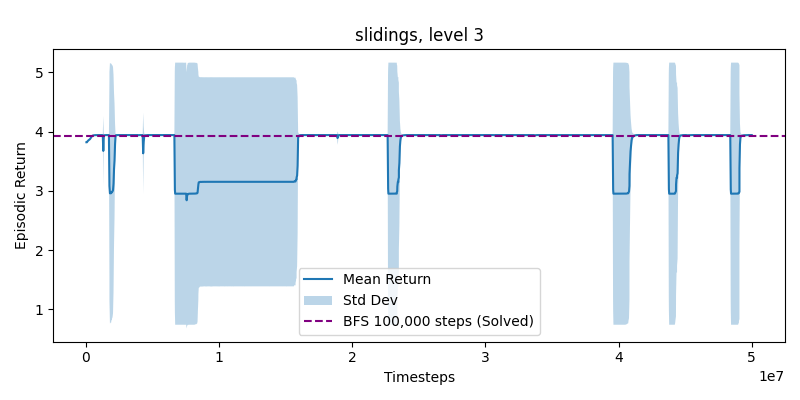}
\end{subfigure}
\begin{subfigure}[t]{0.49\linewidth}
\includegraphics[width=\linewidth]{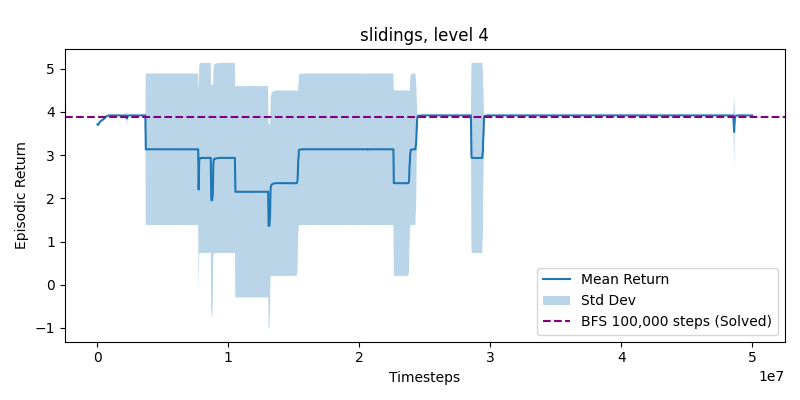}
\end{subfigure}
\begin{subfigure}[t]{0.49\linewidth}
\includegraphics[width=\linewidth]{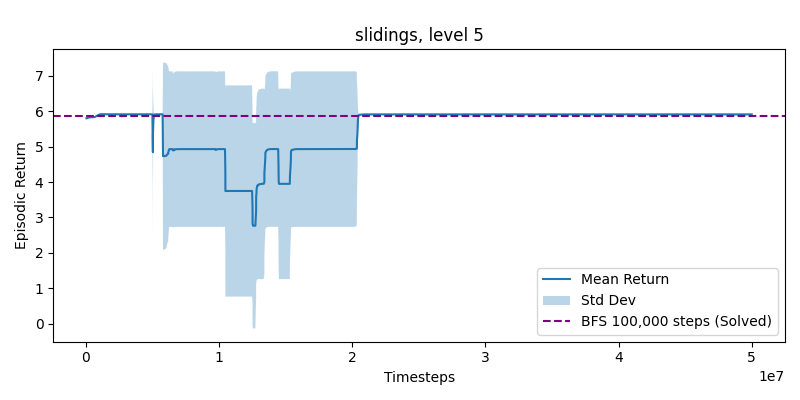}
\end{subfigure}
\begin{subfigure}[t]{0.49\linewidth}
\includegraphics[width=\linewidth]{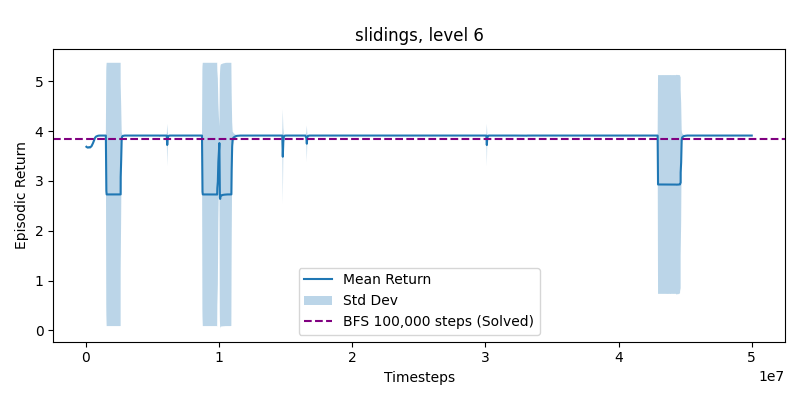}
\end{subfigure}
\begin{subfigure}[t]{0.49\linewidth}
\includegraphics[width=\linewidth]{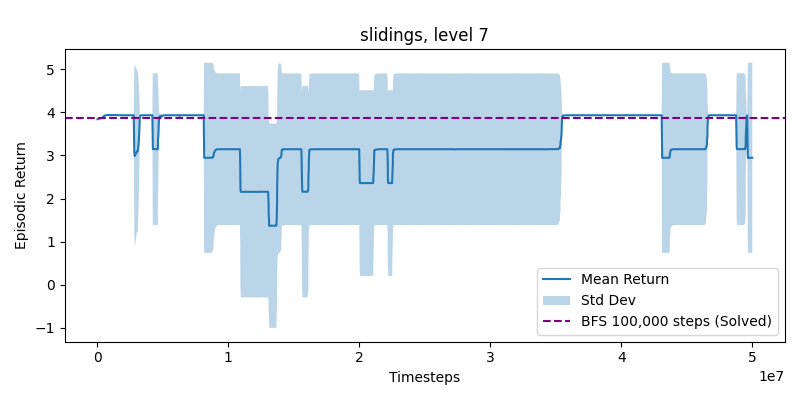}
\end{subfigure}
\begin{subfigure}[t]{0.49\linewidth}
\includegraphics[width=\linewidth]{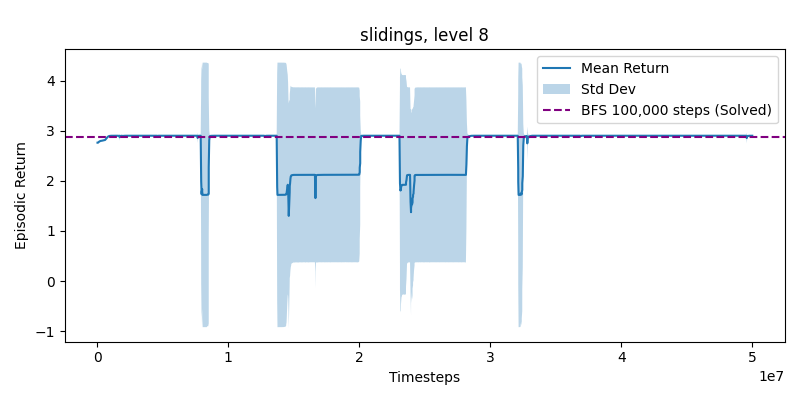}
\end{subfigure}
\begin{subfigure}[t]{0.49\linewidth}
\includegraphics[width=\linewidth]{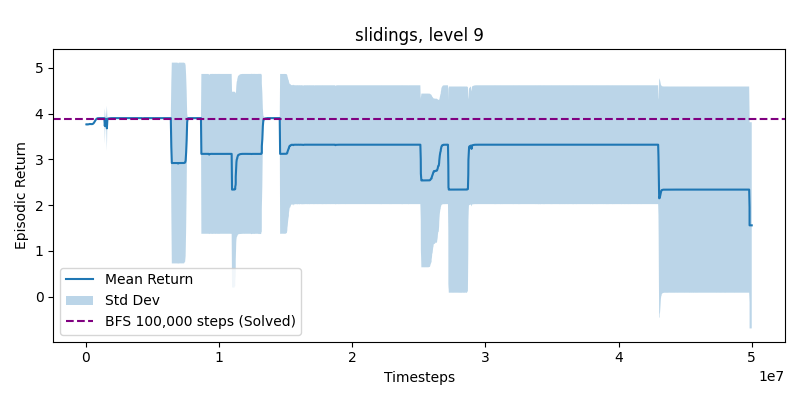}
\end{subfigure}
\begin{subfigure}[t]{0.49\linewidth}
\includegraphics[width=\linewidth]{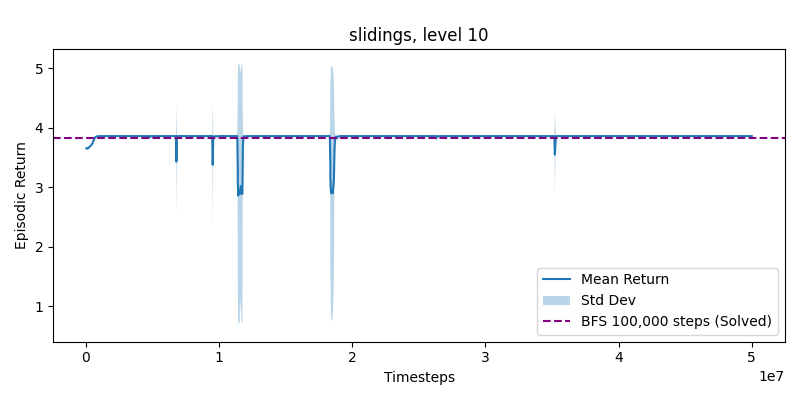}
\end{subfigure}
\caption{Comparison of RL against breadth-first search in \textit{Slidings}. Here, RL rapidly discovers the solutions found via search, though it often exhibits some instability later in training.
% Episode rollouts from RL are pictured on the right.
}
\label{fig:rl_slidings}
\end{figure}

\begin{figure}[t]
\centering
\begin{subfigure}[t]{0.49\linewidth}
\includegraphics[width=\linewidth]{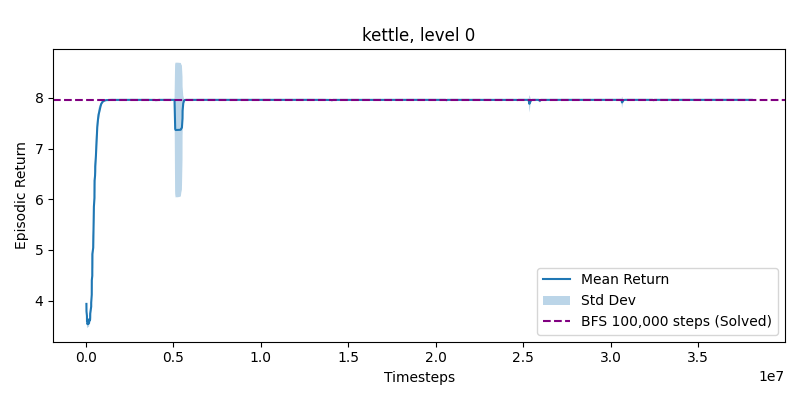}
\end{subfigure}
\begin{subfigure}[t]{0.49\linewidth}
\includegraphics[width=\linewidth]{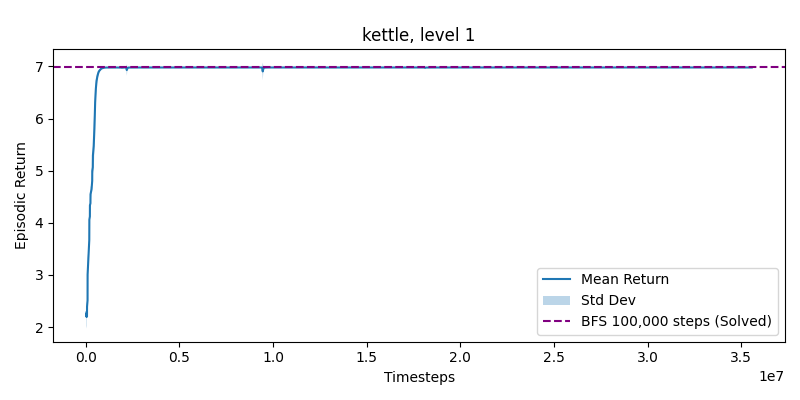}
\end{subfigure}
\begin{subfigure}[t]{0.49\linewidth}
\includegraphics[width=\linewidth]{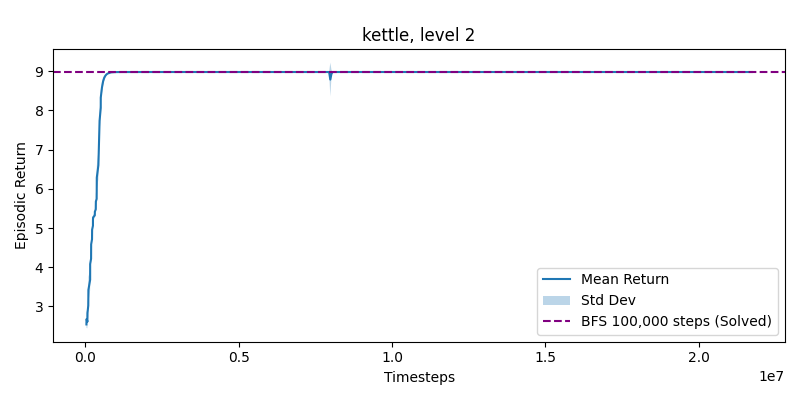}
\end{subfigure}
\begin{subfigure}[t]{0.49\linewidth}
\includegraphics[width=\linewidth]{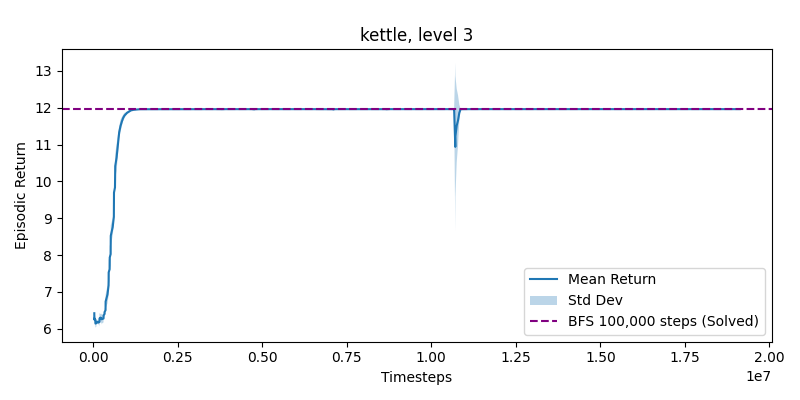}
\end{subfigure}
\begin{subfigure}[t]{0.49\linewidth}
\includegraphics[width=\linewidth]{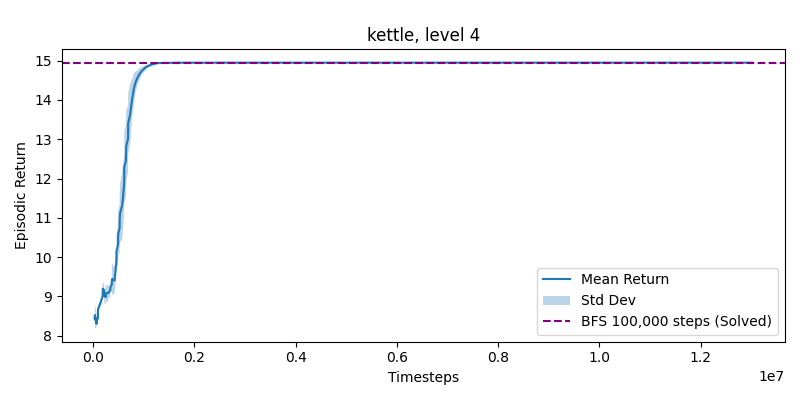}
\end{subfigure}
\begin{subfigure}[t]{0.49\linewidth}
\includegraphics[width=\linewidth]{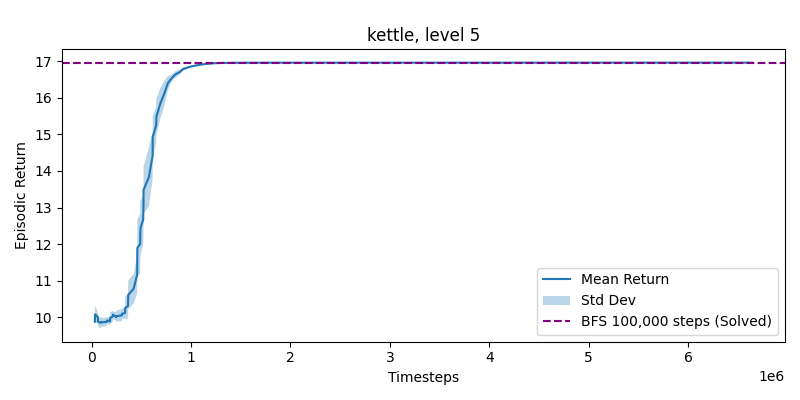}
\end{subfigure}
\begin{subfigure}[t]{0.49\linewidth}
\includegraphics[width=\linewidth]{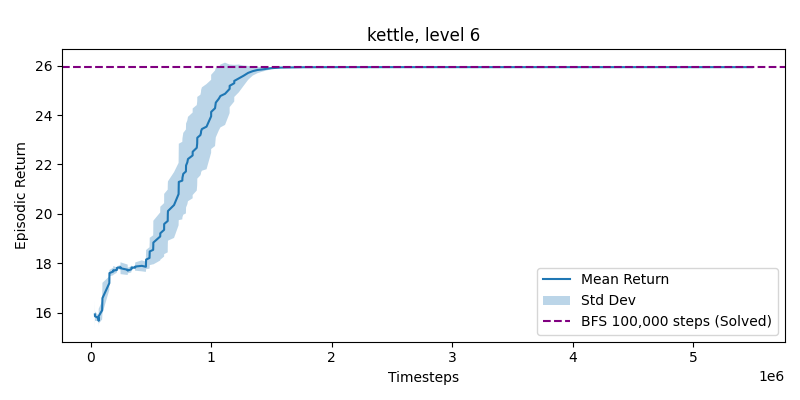}
\end{subfigure}
\begin{subfigure}[t]{0.49\linewidth}
\includegraphics[width=\linewidth]{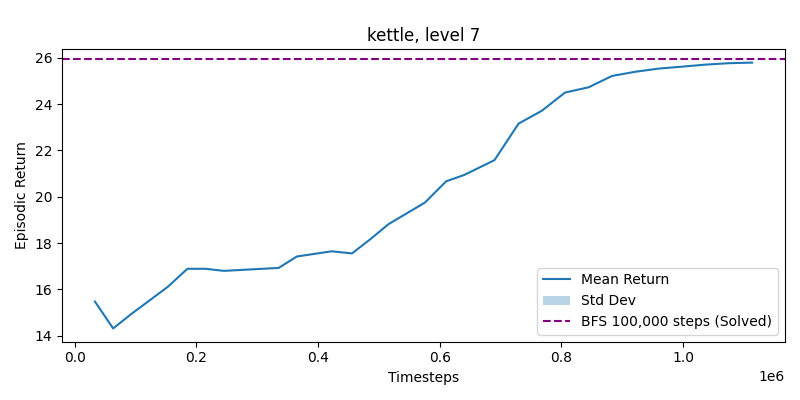}
\end{subfigure}
\caption{Comparison of RL against breadth-first search in \textit{Kettle}. RL agents are able to find optimal solutions, which involve a short sequence of actions, though the time taken to learn this optimal strategy steadily increases as levels (and optimal action sequences) grow and complexify.
}
\label{fig:rl_kettle}
\end{figure}

\begin{figure}[t]
\centering
\begin{subfigure}[t]{0.49\linewidth}
\includegraphics[width=\linewidth]{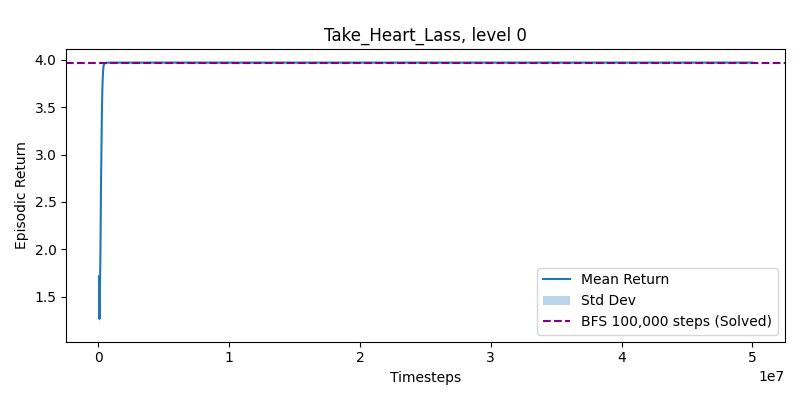}
\end{subfigure}
\begin{subfigure}[t]{0.49\linewidth}
\includegraphics[width=\linewidth]{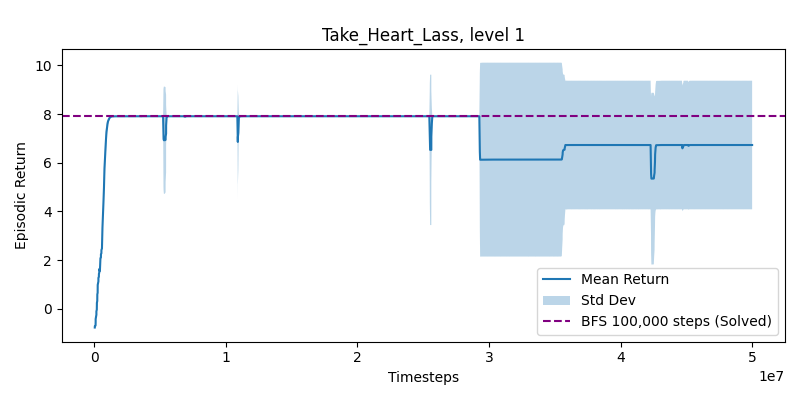}
\end{subfigure}
\begin{subfigure}[t]{0.49\linewidth}
\includegraphics[width=\linewidth]{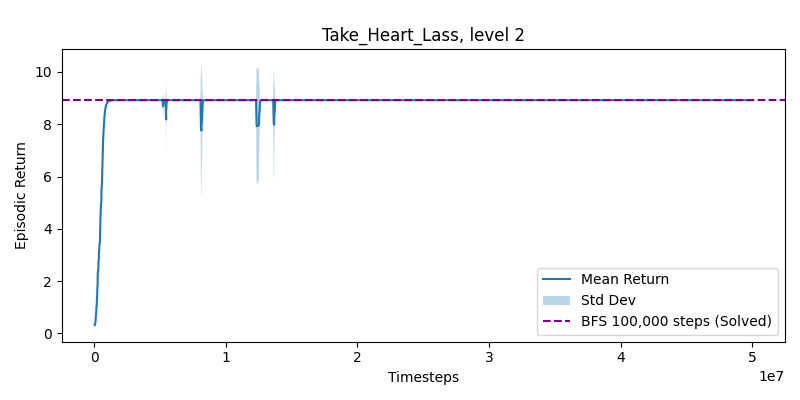}
\end{subfigure}
\begin{subfigure}[t]{0.49\linewidth}
\includegraphics[width=\linewidth]{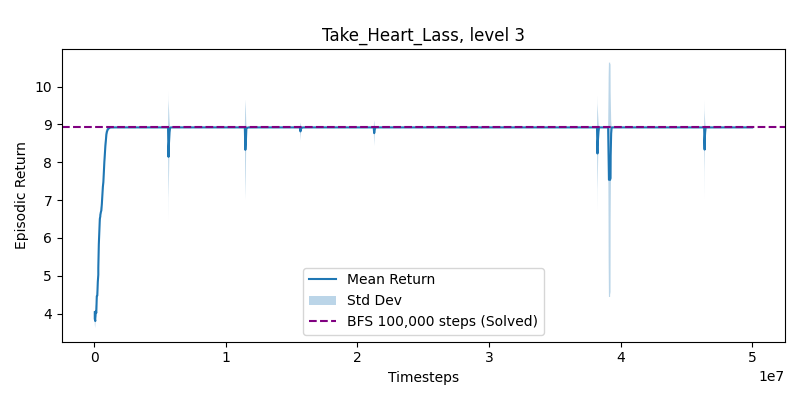}
\end{subfigure}
\begin{subfigure}[t]{0.49\linewidth}
\includegraphics[width=\linewidth]{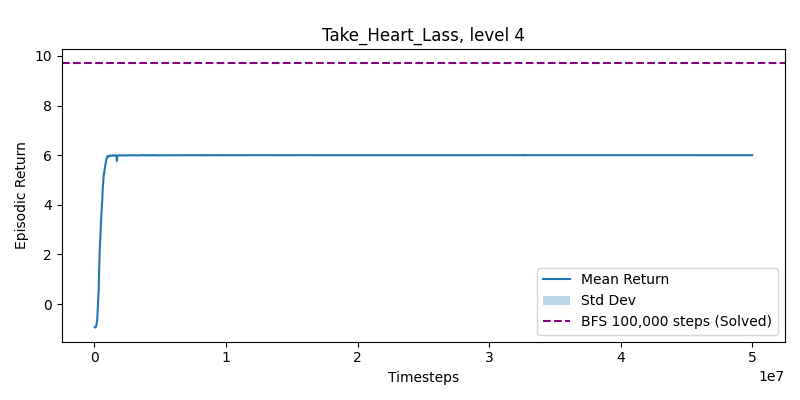}
\end{subfigure}
\begin{subfigure}[t]{0.49\linewidth}
\includegraphics[width=\linewidth]{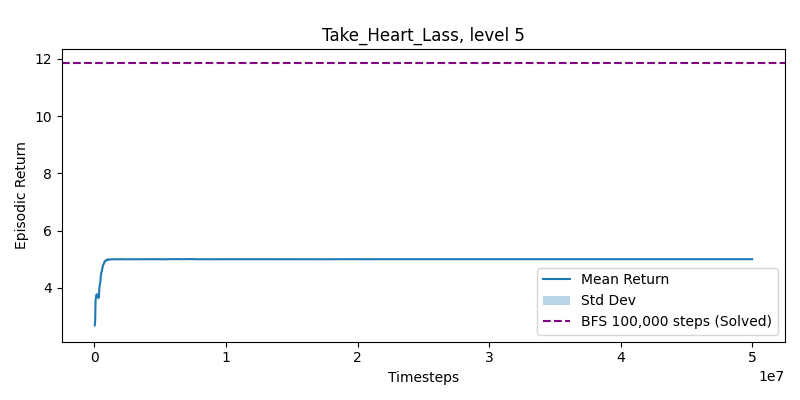}
\end{subfigure}
\begin{subfigure}[t]{0.49\linewidth}
\includegraphics[width=\linewidth]{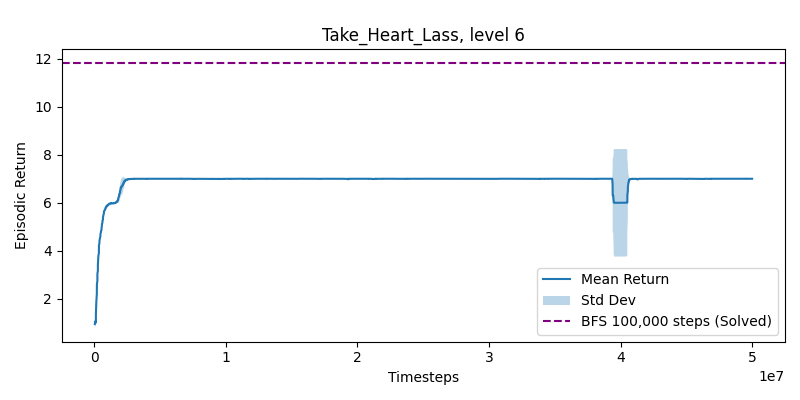}
\end{subfigure}
\begin{subfigure}[t]{0.49\linewidth}
\includegraphics[width=\linewidth]{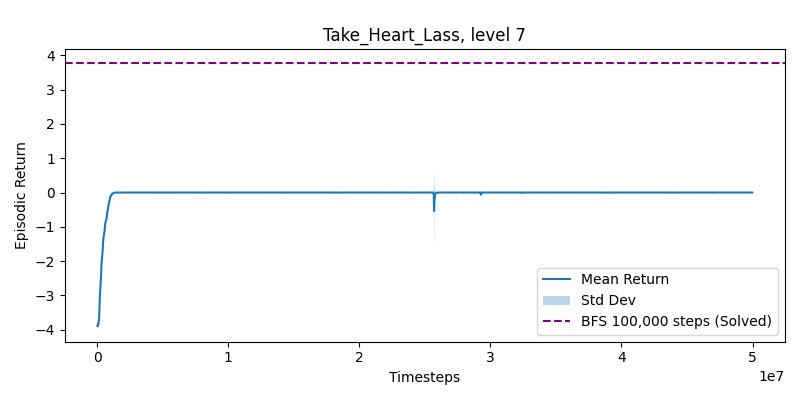}
\end{subfigure}
\caption{Comparison of RL against breadth-first search in \textit{Take Heart Lass}.
RL can handily find solutions to early levels which involve effectively evolve running away from encroaching despair and toward a goal, but it has difficulty in later levels that introduce the use of pushable hearts to strategically block the despair's advance.
}
\label{fig:rl_lass}
\end{figure}

While RL can be deceived by the heuristic functions of \textit{Sokoban Basic} (\autoref{fig:rl_sokoban}) and \textit{Limerick}~(\autoref{fig:rl_limerick}), in which positive reward can be sparse and optimal solutions may require first moving circuitously ``away'' from rewarding states, it does well in games admitting very short solutions such as \textit{Slidings} (\autoref{fig:rl_slidings}) and \textit{Kettle} (\autoref{fig:rl_kettle}), and games that constitute dense reward combinatorial optimization problems such as \textit{Notsnake} (\autoref{fig:rl_notsnake}), where it even discovers a better solution than did breadth-first search after 1 million environment steps (though it still does not discover the \textit{exact} solution). (Note however that this does not necessarily constitute a fair comparison, which would arguably require running search for an equal number of environment steps, and/or comparing the wall clock times of each algorithm.)
In \textit{Take Heart Lass} (\autoref{fig:rl_lass}), agents perform well in early levels which effectively constitute a simple control task involving running away from the encroaching despair and toward a goal, whereas on later levels that require efficiently pushing blocks to clear paths in the knick of time, or block or undo the propagation of despair tiles, the agent often runs into dead-ends or otherwise winds up trapped by despair tiles while attempting to bee-line toward to goal tile.

\begin{figure}
\centering
\begin{subfigure}[t]{.49\linewidth}
\includegraphics[width=\linewidth, trim=20 0 0 20, clip]{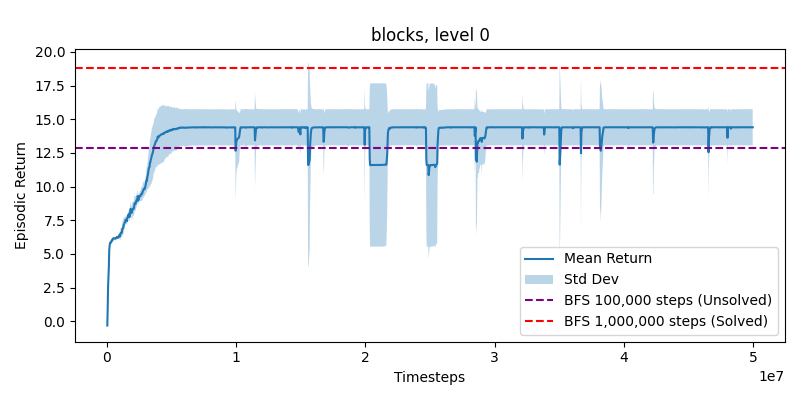}
\end{subfigure}
\begin{subfigure}[t]{.49\linewidth}
\raisebox{9 mm}{
\includegraphics[width=.30\linewidth]{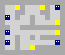}
\includegraphics[width=.30\linewidth]{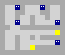}
\includegraphics[width=.30\linewidth]{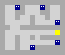}
}
\end{subfigure}
\caption{In \textit{Blocks}, a PPO Reinforcement Learning agent quickly learns to increase score according to the heuristic, but falls into a sub-optimal strategy in which one of the Player blocks is trapped in a dead-end corridor adjacent to the last remaining target.}
\label{fig:rl_curve}
\end{figure}

\begin{figure}
\centering
\begin{subfigure}[t]{0.49\linewidth}
\includegraphics[width=\linewidth]{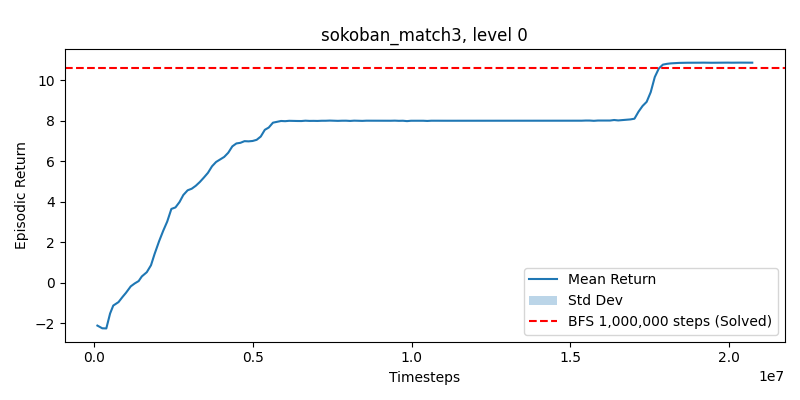}
\end{subfigure}
% \begin{subfigure}[t]{0.49\linewidth}
% \raisebox{9 mm}{
% \includegraphics[width=0.3\linewidth]{figs/sokoban_level-0/im-0.png}
% \includegraphics[width=0.3\linewidth]{figs/sokoban_level-0/im-3.png}
% \includegraphics[width=0.3\linewidth]{figs/sokoban_level-0/im-9.png}
% }
% \end{subfigure}
\begin{subfigure}[t]{0.49\linewidth}
\includegraphics[width=\linewidth]{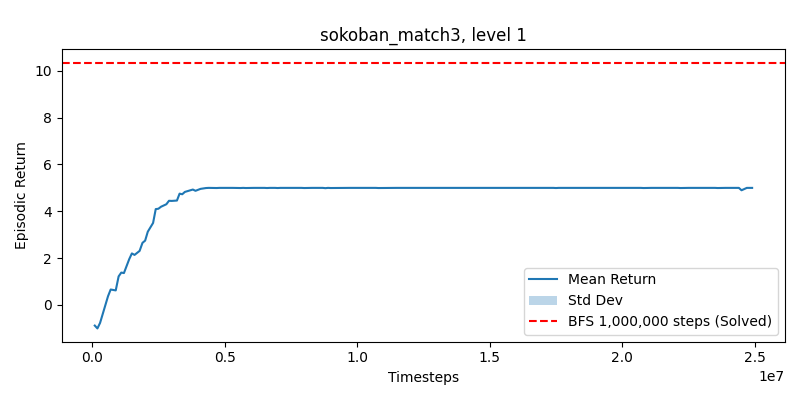}
\end{subfigure}
% \begin{subfigure}[t]{0.49\linewidth}
% \raisebox{9 mm}{
% \includegraphics[width=0.3\linewidth]{figs/sokoban_level-1/im-0.png}
% \includegraphics[width=0.3\linewidth]{figs/sokoban_level-1/im-12.png}
% \includegraphics[width=0.3\linewidth]{figs/sokoban_level-1/im-13.png}
% }
% \end{subfigure}
\caption{Comparison of RL against breadth-first search in \textit{Sokoban Match 3}.
% Episode rollouts from RL are pictured on the right.
}
\label{fig:rl_sokoban_match3}
\end{figure}

\begin{figure}[t]
\centering
\begin{subfigure}[t]{0.49\linewidth}
\includegraphics[width=\linewidth]{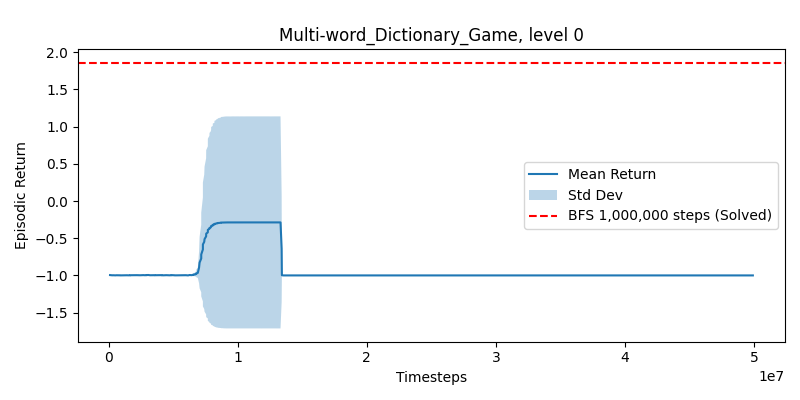}
\end{subfigure}
\begin{subfigure}[t]{0.49\linewidth}
\raisebox{9 mm}{
\includegraphics[width=0.3\linewidth]{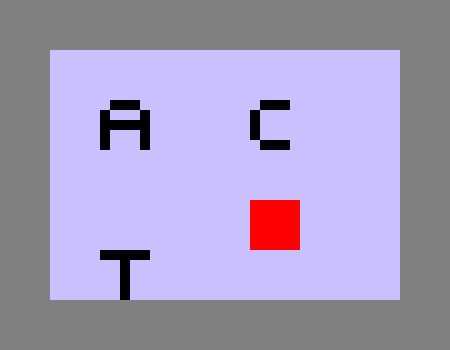}
\includegraphics[width=0.3\linewidth]{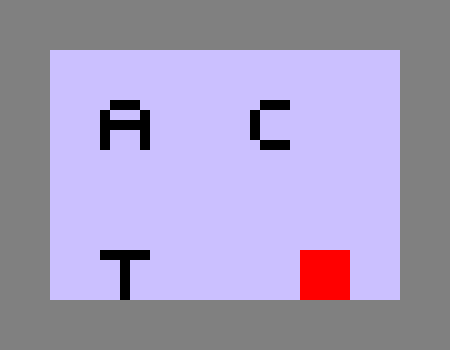}
\includegraphics[width=0.3\linewidth]{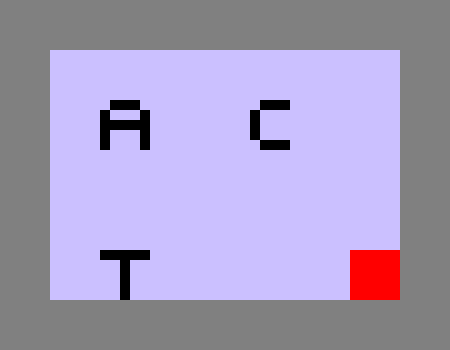}
}
\end{subfigure}
\caption{Comparison of RL against breadth-first search in \textit{Multi-Word Dictionary Game}.}
\label{fig:rl_multi}
\end{figure}

% \subsection{Tree search}

% We run breadth-first search on our full dataset of games, capping the number of environment steps at $100,000$, and setting a timeout of 1 minute in \autoref{tab:bfs_results}.

\clearpage
\begin{lstlisting}[style=puzzlescript,frame=none,backgroundcolor=\color{ps-background},caption={Example of a \puzzlescript file (\textit{LimeRick})}]
title Lime Rick
author Tommi Tuovinen
homepage http://www.kissmaj7.com/

(Ported with the very gracious permission of Tommi Touvinen
The first ten levels of a neato game - you can play the full version here
http://www.kongregate.com/games/KissMaj7/lime-rick
The full version includes some mechanics that aren't covered in the levels here,
but they are supported.)

========
OBJECTS
========

Background
black

Exit
red
.000.
00000
00000
00000
.000.

Apple
blue
.000.
00000
00000
00000
.000.

PlayerBodyH
green
.000.
00000
0...0
00000
.000.

PlayerBodyV
green
.000.
00.00
00.00
00.00
.000.

Crate
orange
00000
0...0
0...0
0...0
00000

PlayerHead1
lightgreen
.000.
0.0.0
00000
00000
.000.

PlayerHead2
yellow
.000.
0.0.0
00000
00000
.000.

PlayerHead3
orange
.000.
0.0.0
00000
00000
.000.

PlayerHead4
red
.000.
0.0.0
00000
00000
.000.

Wall
brown

=======
LEGEND
=======

Player = PlayerHead1 or PlayerHead2 or PlayerHead3 or PlayerHead4
Obstacle = PlayerBodyH or PlayerBodyV or Wall or Crate or Player
PlayerBody = PlayerBodyH or PlayerBodyV
. = Background
P = PlayerHead1
# = Wall
E = Exit
A = Apple
C = Crate

=========
SOUNDS
=========

sfx0 3295707 (player jump)
sfx1 3538707 (player jump to max)
sfx2 42451307 (player move horizontally)
endlevel 96434300
startgame 49875902

================
COLLISIONLAYERS
================

Background
Exit, Apple
PlayerBody
Player, Wall, Crate

======
RULES
======

(this game handles all the movement stuff itself - it removes all movements before 
the movement phase has a chance to tick at all)

UP [ UP PlayerHead4 ] -> [ PlayerHead4 ] 
UP [ UP PlayerHead3 | No Obstacle ] -> [ PlayerBodyV | PlayerHead4 ] sfx1
UP [ UP PlayerHead2 | No Obstacle ] -> [ PlayerBodyV | PlayerHead3 ] sfx0
UP [ UP PlayerHead1 | No Obstacle ] -> [ PlayerBodyV | PlayerHead2 ] sfx0

horizontal [ > Player | Crate | No Obstacle ] -> 
               [ PlayerBodyH | PlayerHead1 | Crate ] sfx2

horizontal [ > Player | No Obstacle ] -> [ PlayerBodyH | PlayerHead1 ] sfx2

[ Player Apple ] [ PlayerBody ] -> [ Player Apple ] [ ] 
[ Player Apple ] -> [ Player ] 

[ > Player ] -> [ Player ] 

DOWN [ Player | No Obstacle ] -> [ PlayerBodyV | PlayerHead1 ] 
DOWN [ Crate | No Obstacle ] -> [ | Crate ] 

==============
WINCONDITIONS
==============

some player on exit

=======
LEVELS
=======

message level 1 of 10

###################
#.................#
#.................#
#............#....#
#............#....#
#.......#...##....#
#..P....#...##..E.#
###################
###################
..#...#...#...#...#
#...#...#...#...#..
###################
###################
###################
###################

(additional levels omitted for clarity)
message congratulations!
\end{lstlisting}

% \newpage
% \input{tables/valid_games}

% \newpage
% \input{tables/partial_valid_games}

% \newpage
% \input{tables/bfs_results}

% \bibliographystylesec{plain}
% \bibliographysec{refs_appendix}

\section{Use of Large Language Models in Paper Writing}

Large Language Models were used to help refine the grammar in initial drafts of our section on the results of experiments using LLMs to play games in \name. All text was thoroughly reviewed by human authors prior to submission.

\end{document}